\documentclass[sigconf, 10pt, nonacm, letterpaper]{acmart}

\usepackage{array}
\usepackage{colortbl}
\usepackage{xurl}
\usepackage{color}
\usepackage{soul}
\usepackage{siunitx}
\usepackage{listings}
\usepackage{subcaption}
\usepackage{float}
\usepackage[normalem]{ulem}
\usepackage{setspace}
\usepackage{placeins}
\usepackage{afterpage}
\usepackage{hhline}
\usepackage{enumitem}
\usepackage{pifont}
\usepackage{arydshln}
\usepackage{adjustbox}
\usepackage{lipsum}
\usepackage{tabularx}
\usepackage{makecell}
\usepackage{wrapfig}
\usepackage{moreverb}
\usepackage{multirow}
\usepackage{xspace}
\usepackage{hyperref}
\usepackage{epsfig}
\usepackage{endnotes}
\usepackage{url}
\usepackage{fancyhdr}
\usepackage{cascadia-code}
\usepackage{xcolor}
\usepackage{caption}
\usepackage[framemethod=TikZ]{mdframed}
\definecolor{codegray}{RGB}{128, 128, 128}
\definecolor{backcolor}{RGB}{250, 250, 250}
\definecolor{rulecolor}{RGB}{120, 120, 120}
\definecolor{keywordcolor}{RGB}{218, 75, 88}
\definecolor{stringcolor}{RGB}{64, 135, 39}
\definecolor{identifiercolor}{RGB}{50, 50, 50}

\usepackage{booktabs}

\usepackage{DejaVuSansMono}
\lstdefinestyle{mystyle} {
    commentstyle=\color{rulecolor},
    keywordstyle=\color{keywordcolor},
    stringstyle=\color{stringcolor},
    basicstyle=\ttfamily\scriptsize,
    identifierstyle=\color{identifiercolor},
    backgroundcolor=\color{backcolor},
    breakatwhitespace=false,
    breakindent=0pt,
    breaklines=true,
    captionpos=b,
    keepspaces=true,
    numbers=left,
    numberstyle=\ttfamily\tiny\color{rulecolor},
    numbersep=3pt,
    showspaces=false,
    showstringspaces=false,
    showtabs=false,
    framexleftmargin=0pt,
    frame=lines,
    rulecolor=\color{rulecolor},
    rulesepcolor=\color{gray},
    xleftmargin=0pt,
    xrightmargin=0pt,
}
\definecolor{blueframecolor}{RGB}{173,212,251} 
\definecolor{textgray}{gray}{0.5}
\definecolor{typeflyblue}{RGB}{95,148,247} 
\definecolor{typeflyred}{RGB}{202,85,92}
\definecolor{typeflygreen}{RGB}{82,133,54}
\definecolor{typeflypurple}{RGB}{189,21,240}

\newmdenv[  
  linecolor=blueframecolor,
  outerlinewidth=0.5pt, 
  roundcorner=3pt, 
  backgroundcolor=blueframecolor!10,
  frametitlerule=true,
  innertopmargin=3pt,
  innerbottommargin=3pt,
  font=\small
]{customframe}

\definecolor{lightgray}{gray}{0.9}
\definecolor{lightblue}{rgb}{0.9,0.9,1}
\definecolor{blue_bg}{rgb}{0.85,0.85,1}
\definecolor{lightyellow}{rgb}{1,1,0.8}
\definecolor{lightpurple}{rgb}{1,0.85,1}
\definecolor{red}{rgb}{1,0,0}
\definecolor{darkgreen}{rgb}{0.4,0.7,0.3}
\definecolor{lightcyan}{rgb}{0.4, 0.8, 0.9}

\usepackage{tcolorbox}

\newcommand{\remove}[1]{}

\newcommand\rt{time-sensitive\xspace}

\newcommand{\myparagraph}[1]{\vspace{2mm} \noindent \textit{\textsf{#1}}}
\newcommand{\skill}[1]{\textsl{`#1'}}

\newcommand\sysname{\textsc{TimelyLLM}\xspace}

\newcommand\datasetname{LRTrace\xspace}

\title{\sysname: Segmented LLM Serving System for Time-sensitive Robotic Applications}

\author{Neiwen Ling}
\affiliation{\institution{Yale University} 
\country{}}
\email{neiwen.ling@yale.edu}

\author{Guojun Chen}
\affiliation{\institution{Yale University} 
\country{}}
\email{guojun.chen@yale.edu}

\author{Lin Zhong }
\affiliation{\institution{Yale University} 
\country{}}
\email{lin.zhong@yale.edu}

\begin{document}
\begin{abstract}
Large Language Models (LLMs) such as GPT-4 and Llama3 can already comprehend complex commands and process diverse tasks. This advancement facilitates their application in controlling drones and robots for various tasks. However, existing LLM serving systems typically employ a first-come, first-served (FCFS) batching mechanism, which fails to address the time-sensitive requirements of robotic applications. To address it, this paper proposes a new system named \sysname serving multiple robotic agents with time-sensitive requests. \sysname introduces novel mechanisms of segmented generation and scheduling that optimally leverage redundancy between robot plan generation and execution phases. We report an implementation of \sysname on a widely-used LLM serving framework and evaluate it on a range of robotic applications. Our evaluation shows that \sysname improves the time utility up to $1.97\times$, and reduces the overall waiting time by $84\%$.

\end{abstract}


\maketitle

\section{Introduction}
\label{sec:intro}
Large Language Models (LLMs) have opened new directions for robotic applications, particularly in Embodied Artificial Intelligence (AI), enabling the execution of complex tasks in environments that were previously impossible~ \cite{wang2024llm, rana2023sayplan, ren2023robots, chen2023typefly, zhang2023bootstrap}. 
These models have been integrated into systems for commercial robots, including the Figure Humanoid Robot~\cite{figurerobot} and Boston Dynamics Spot Robot Dog~\cite{bostondynamics}.
We envision a future in which a large number of robotic agents continuously and concurrently send requests to an LLM service  \cite{wang2024llm, chen2023typefly, achiam2023gpt, vemprala2024chatgpt}, with \rt requirements. They require the LLM service to process a request and return responses within a deadline, and the utility of the service drops as the response passes the deadline. This is essential for robotic agents to react in time, maintaining synchronization with their environments and avoiding potential task failures. 

Unfortunately, the modern LLM engines~\cite{kwon2023efficient, agrawal2024taming,tensorrt_llm, yu2022orca} employ the FCFS batching mechanism to serve multiple requests and as a result, do not adequately address the \rt requirements of robotic requests~\cite{seakhoa2019revenue, barcis2019evaluation, pelikan2023managing}.
The batching approach prioritizes resource utilization over the latency of individual requests. Furthermore, the FCFS policy is inflexible when handling diverse request priorities. By processing requests in the order of arrival, it cannot distinguish between high- and low-priority tasks, potentially delaying urgent requests when less critical ones arrive first. 

Our key insight is that modern LLM services can often generate a plan much faster than the robotic agent can execute it. For instance, GPT-4o~\cite{hurst2024gpt} can generate close to 100 tokens per second while a drone may take a few seconds to execute a simple command of a few tokens like moving forward.
Thus, there is often significant time redundancy between the robotic plan generation and execution. This presents an opportunity for the temporary suspension of plan generation for robotic agents awaiting completion of the current plan, thereby enabling the reallocation of computational resources towards more urgent requests.

To harness these insights, we developed the \sysname system, which provides \rt LLM serving for concurrent and continuous requests across various agents. Our approach is founded on two pivotal decisions:
(\textit{i}) First, \sysname partitions a long LLM generation procedure into segments. In this context, `long' refers not only to the number of output tokens but also to the extended execution time these outputs entail. For instance, an LLM-generated plan such as \skill{mu(100);mf(50)} might be short in length but require substantial time to execute.
\textit{ii}) Second, \sysname schedules the generation for the segments to comply with the \rt demands of each request. For example, in robotic task planning, the system generates the initial segment within the specific tolerable response time (e.g., 1 second \cite{shneiderman1984response}) and subsequent segments before completing the preceding plan execution. 
In this way, the system effectively utilizes the time redundancy between the robot task plan generation and execution stages to disperse the workload within a time duration, thereby alleviating resource contention and maximizing compliance with the \rt demands of each request.



We implement \sysname on the widely-used LLM serving framework vLLM \cite{kwon2023efficient}. We also build a dataset collection system \datasetname to construct time traces from real-world robots, including the Tello drone \cite{dji2023tellosdk} and Neuromeka robot arm \cite{robotarmindy}, aiding in the evaluation of multi-agent LLM serving systems.
Using this dataset collection system, we evaluate \sysname on different robotics applications across a range of workloads with different task types, varying job parallel degrees, arrival rates, and agent numbers. We also evaluate the design choices, scalability, and overhead of \sysname. Experimental results show that compared to the state-of-the-art vLLM, \sysname achieves up to a $1.97 \times $ improvement in the time utility of robotic task response time, and reduces the overall robot waiting time by $84\%$. 
We summarize the contributions of this paper as follows:
\begin{itemize}[leftmargin=1em]
     \item We propose \sysname, the first time-sensitive LLM serving system for robotic applications, which better meets the time requirements of multiple robots by leveraging the redundancy between LLM plan generation and robot execution.  
     \item We adopt an efficient segmented generation mechanism with a robotic query-specific stop checker and context caches. We also design a segmented scheduling strategy considering both time utility gain and robot execution efficiency, enabling our system to handle multiple robotic requests with optimized time utility.
     \item We design and build \datasetname, the first dataset collection system enabling flexible creation of real-world time traces for multiple LLM-powered robots. Based on it, we evaluate the performance of \sysname through extensive experiments with a range of real workloads.
\end{itemize}


\section{Background}
\label{sec:background}
\subsection{Latency Profile of LLM Service}
\label{subsec:llm_latency}

\begin{table}[t]
\centering
\caption{\textmd{\small LLM Latency for a Single Request Across Different Phases: Tokenization, Decoding, and Detokenization. The table demonstrates that decoding dominates the entire inference procedure. We test Llama3-8B-float16 and Phi-3-mini-128k-float16, implementing them using Hugging Face Transformers \cite{hf_transformer} on an NVIDIA RTX 4090.}}
\label{tab:latency}
\renewcommand{\arraystretch}{0.9} 
{\footnotesize
\begin{tabular}{@{}>{\raggedright\arraybackslash}p{1.9cm}>{\raggedright\arraybackslash}p{1.7cm}>{\raggedright\arraybackslash}p{2cm}>{\raggedright\arraybackslash}p{2cm}@{}}
\toprule
\textbf{Phase}               & \textbf{Tokens} & \textbf{Llama3} & \textbf{Phi3} \\ \midrule
Tokenization               & 2884            & 2.609 ms (0.43\%)        & 1.541 ms (0.18\%)    \\
Decoding 0                 & token 0              & 328.45 ms           & 563.04 ms      \\
Decoding 1                 & token 1              & 21.93 ms            & 22.24 ms                         \\
Decoding 2                 & token 2              & 21.56 ms            & 20.99 ms                        \\
3 to N-1            &  ...             & ...            & ...                       \\
Decoding N             & last token               & 21.77 ms            & 22.00 ms                        \\                
Decoding            & 15             & 611.2 ms (99.57\%)           & 841.7 ms  (99.80\%)   \\
Detokenization                      & 15              & 0.052 ms (0.008\%)       & 0.091 ms (0.011\%)    \\
\textbf{Total time}                      & in:2884, out:15          & 613.82 ms             & 843.36 ms                 \\ \bottomrule
\end{tabular}
}
\end{table}

A typical LLM service processes each incoming request in three stages: tokenization, decoding, and detokenization \cite{brown2020language,llama3}.
\emph{Tokenization} converts text into a sequence of tokens that the model can process. 
\emph{Decoding} predicts each subsequent token based on the tokens generated so far. Finally, \emph{Detokenization} converts the output tokens back into human-readable text. 

Decoding is the most time-consuming stage. \autoref{tab:latency} presents the latency measurement for serving one request with both Meta Llama3 \cite{llama3} and Microsoft Phi-3 \cite{abdin2024phi}. In both cases, the decoding phase accounts for over $99\%$ of the total time.
Because the model generates tokens sequentially, each token prediction requires the entire model to process preceding tokens, causing the latency to grow linearly with respect to the number of output tokens \cite{chen2023typefly}. 
Given the complexity and large size of models like those based on GPT or Llama, the number of parameters involved in each token generation is substantial—often in the billions. For example, GPT-4 involves approximately 1.8 trillion parameters \cite{gpt4_parameters}, while even the smallest version of the Llama 3 model has around 8 billion parameters \cite{meta_llama_3}. This massive computational load further extends the decoding time. 
Decoding the first token (token 0 in \autoref{tab:latency}) is particularly expensive because it requires \emph{Prefilling} that computes the attention states for input tokens. This computation is unnecessary for subsequent tokens due to the use of the KV cache~\cite{pope2023efficiently} and for requests with repeated prompts, thanks to Prompt cache~\cite{gim2024prompt}.

To improve GPU utilization and enhance system throughput, LLM serving engines such as vLLM \cite{vllm} and TensorRT-LLM \cite{tensorrt_llm} commonly adopt batch processing as a strategy.
We build \sysname on top of \emph{continuous batching} \cite{yu2022orca,kwon2023efficient}, which allows a new request to be added for processing as soon as resources become available, providing a more flexible batching mechanism.
However, even with continuous batching, the inherent limitations of batching persist - trade the service latency of individual requests for high resource utilization. 
Moreover, most batching systems \cite{kwon2023efficient, tensorrt_llm, yu2022orca, agrawal2024taming} employ the FCFS scheduling policy that ignores the urgency of requests. 
For instance, if a non-urgent request arrives before a more time-sensitive one, the FCFS policy would still process the former first, leaving fewer resources available for the latter. This indiscriminate approach to task queueing can hinder the system's ability to meet specific \rt demands.
To effectively address these challenges, our system \sysname introduces a novel LLM scheduling mechanism. This approach prioritizes tasks according to their urgency and strategically batches requests according to their specific \rt requirements. 

\subsection{LLM-powered Robotics and its Latency}
\label{subsec:robot_latency}
In LLM-powered robot control, robots leverage the LLM's reasoning capabilities for planning a task given by the user instructions~\cite{wang2024llm, chen2023typefly} or handling unexpected situations\cite{liu2023reflect, skreta2023errors}. 
Due to the use of in-context learning, robotic requests to the LLM are usually very long, including the user's instructions, information about the world, robot skill sets, guidance for plan generation, and case examples.
The output is typically a sequence of robot actions, e.g., \skill{pick(cyan\_box)->place(cyan\_box)->
pick(red\_box)} for a robot arm \cite{wang2024llm}. Notably, the robot can start executing the planned actions, such as pick(cyan\_box), while subsequent steps place and pick are still under generation. This provides an opportunity for LLM to generate robot plans in a more flexible way, which will be reflected in our design.

For LLM-based robotic planning, two latencies of the LLM service are important to the end user and application~\cite{chen2023typefly}. The first, called \emph{response time}, is the time from when the service receives a task to when the robot starts to act.  The second, called \emph{task completion time}, is the time from when the service receives a task to when the robot completes the task. 
This metric comprises two parts: robot waiting time, defined as the total latency introduced by the LLM planning, and robot execution time, which depends on the content of the task plan.

Most existing LLM services generate an entire plan before robotic execution and as a result, suffer from long response and task completion time. Typefly, a recent system~\cite{chen2023typefly}, allows robotic execution before the complete plan is available, via stream interpretation. It reduces both response and task completion time. However, it generates a plan nonstop and as a result, does not reallocate resources between concurrent generations or accommodate different \rt requirements in response time for different types of tasks in robot control~\cite{barcis2019evaluation, seakhoa2019revenue, ravindran2005tuf, tidwell2010tuf}.

\sysname reduces the response time without increasing the task completion time while considering the \rt requirements. 

\subsection{TUF for LLM serving in Robotics}
\label{subsec:app_llm}
Robotic tasks are usually time-critical: the system yield depends on its ability to complete tasks within the required time constraints. The \emph{time-utility function} (TUF)~\cite{jensen1985time} specifies the utility from serving a task as a function of latency. It allows the developer/user to express diverse timing requirements. Different robotic applications, as well as tasks within the same application, may have varying TUFs.
\sysname aims to serve a large number of LLM requests concurrently in a time-sensitive manner, optimizing their aggregated TUFs.

\sysname considers a general form of TUF shown in \autoref{eq:tuf_robot}, as extended from a real-world case example in~\cite{maynard1988example}. We adjust its parameters to accommodate different robotic tasks. 
\begin{equation}
TUF(t) = \min(\beta, \alpha(t - ERT)+\beta)
\label{eq:tuf_robot}
\end{equation}

\noindent where $t$ is the response time,
$ERT$ the expected response time or deadline, $\alpha$ and $\beta$ the utility decline slope and the initial time utility, respectively. All are specified by the user or the developer to represent their \rt requirements.
In scenarios for normal robot tasks such as task planning based on the user's instruction, slight delays are more permissible, allowing a robot to process complex queries. Previous research in human-computer interaction (HCI) indicates a user preference for delays not exceeding one second \cite{shneiderman1984response}, so we set the ERT for normal tasks to 1 second. In contrast, for urgent tasks such as error handling, e.g., when a human approaches the robot rapidly, quicker responses are critical. Past research on drone obstacle avoidance~\cite{carrio2018drone} indicates that a reaction time of 200 ms is necessary for a drone to fly at 2 m/s. Accordingly, we set the ERT for urgent tasks such as error handling to 200 ms. 
The consequence of missing a deadline on urgent tasks is less tolerable, resulting in a steeper decline in utility after deadline, i.e., a larger $\alpha$, compared to normal tasks. Furthermore, the initial utility value assigned to urgent tasks ($\beta$) is higher, reflecting its critical nature to ensure a timely response. From the perspective of LLM serving systems, multiple levels of $\beta$ can be defined, allowing users (robotic agents) to self-subscribe to the appropriate urgency level based on their specific task expectations.

For \rt LLM serving, it is essential to prioritize resource allocation to tasks that maximize overall time utility. 
Current LLM serving systems often operate under the implicit assumption that token generation for one robot action will immediately follow the previous robot action without interruption, hindering in-time preemption to optimize resource allocation. 
From an application perspective, the LLM agent continuously receives the generated plan, even when immediate execution is not required.
In this paper, we challenge this assumption by proposing a segmented generation strategy. This strategy enables the suspension of generation for tasks that have already formulated partial plans and require a designated period for execution, allowing the LLM serving system to reallocate resources to other tasks. Our system design is fundamentally centered around this concept.

\section{Overview of \sysname\label{sec:sys_design}}


\sysname is a \rt LLM serving system for concurrent and continuous robot queries. 
The key idea of \sysname is to dynamically schedule the LLM generation (decoding) procedure according to the content of the generated response. It applies this key idea in two stages: (\textit{i})  generating the response in multiple segments tailored to the executability of each segment, or \emph{segmented generation} (\textit{ii}) assigning scheduling priorities to the suspended generation based on the estimated execution time of the generated content.
This design of segmented generation and scheduling enables effective utilization of the time redundancy between the robot task plan generation and execution stages, allocating the computing resources to more urgent requests and ensuring better compliance with the \rt demands of each request.

\begin{figure}[tb]
\centering
\includegraphics[width=0.48\textwidth]{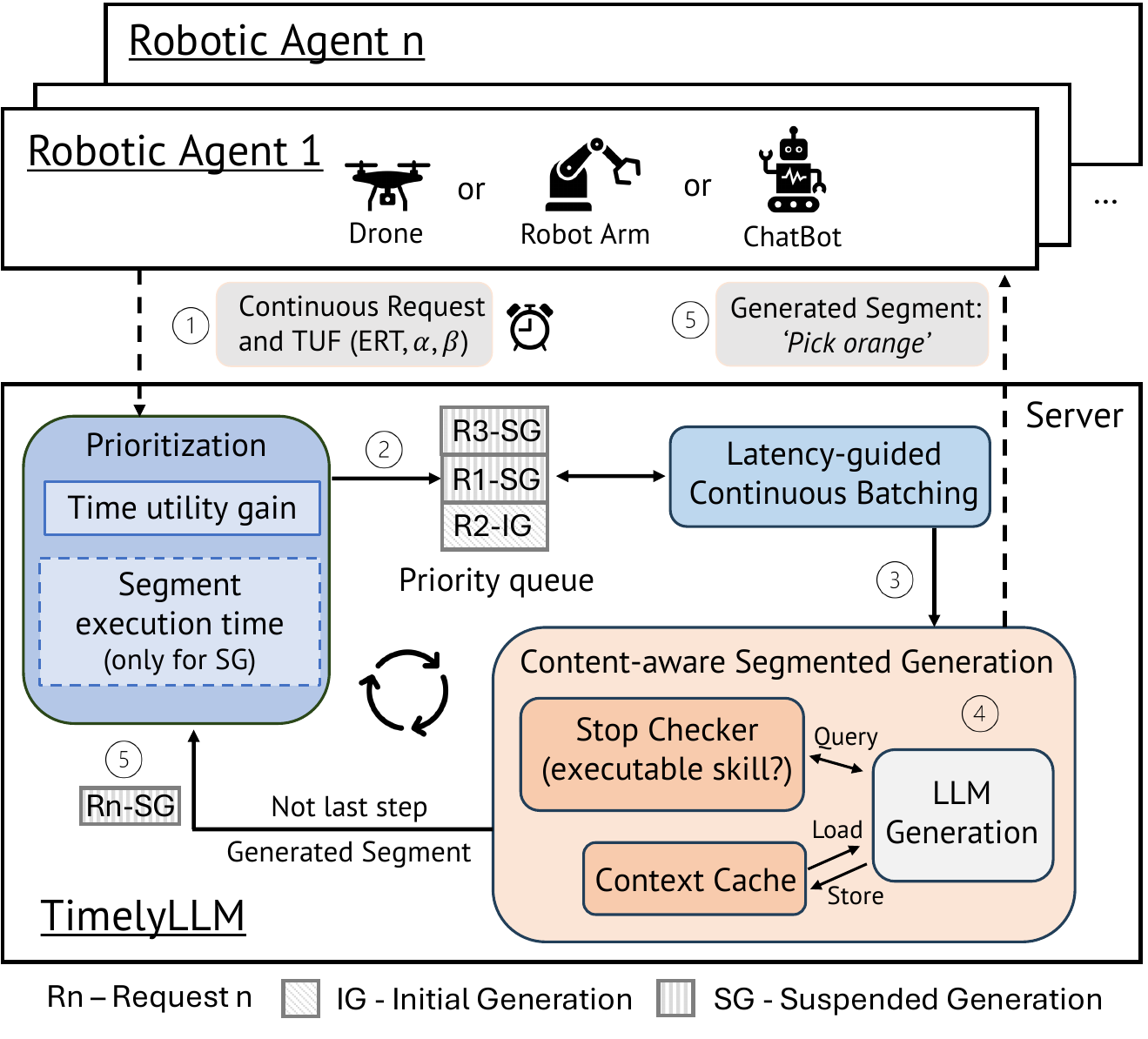}
\caption{\textmd{\small System overview of \sysname: A robotic agent submits a request with time-sensitive requirement defined by TUF (\S\ref{subsec:app_llm}) to \sysname. \sysname generates the response for incoming requests in segments by continuously suspending and resuming the generation (\S\ref{sec:con_infer}). The scheduler in \sysname (\S\ref{sec:seg_schedule}) manages all the initial and suspended generation and (\S\ref{sec:priority_assign}) prioritizes them based on the potential time utility gain and estimated execution time. Additionally, \sysname adaptively adjusts batch sizes to efficiently harness GPU parallel computing resources (\S\ref{sec:batching}).}}
\label{fig:sys_arch}
\end{figure}

\autoref{fig:sys_arch} presents a high-level system overview of \sysname. A robotic agent submits a request, along with its \rt requirement in terms of the Expected Response Time or deadline (ERT in TUF), the tolerance level for missing deadlines ($\alpha$ in TUF), and the time-sensitive degree ($\beta$ in TUF). We encapsulate these parameters in the TUF defined in \S\ref{subsec:app_llm} to form the basis of the \rt requirement.

\sysname uses a priority queue for generations. When a new request arrives, \sysname assigns a priority to its generation and places the generation in the queue. 
 \sysname evaluates resource availability whenever the LLM inference engine completes a decoding iteration, and when resources permit, it removes the highest-priority generation from the queue and continues with the LLM inference engine.

\sysname implements \emph{segmented generation}, where the response is generated in segments rather than all at once. Specifically, \sysname continuously monitors the generated content; when it identifies some executable statement(s) in the newly generated response, \sysname suspends the generation and places it into the priority queue. To ensure timely execution, \sysname also dispatches the generated segment to the corresponding agent for immediate action. 
A key challenge of segmented generation lies in minimizing the overhead associated with resuming LLM generation. We address this issue by carefully designing the cached information for resumption (\S\ref{sec:con_infer}). 

The scheduler then determines the time to resume generation, typically before the completion of the previous segment's execution.
Notably, with segmented generation, \sysname must schedule both initial and suspended generations. 
We formulate the scheduling problem to be compatible with segmented generation (\S\ref{sec:seg_schedule}) and incorporate a priority assignment strategy guided by the potential gain in time utility.
For suspended generations, \sysname also factors in the estimated robot execution time of the generated plan, ensuring resumption urgency aligns with the estimated completion of prior segments (\S\ref{sec:priority_assign}).
To further leverage GPU parallel computing resources, we implement continuous batching. However, this introduces the challenge of batching without influencing time utility in the current generation. We address this issue by carefully adjusting the batch size based on both the time-sensitive requirements of in-process requests and the availability of GPU resources (\S\ref{sec:batching}).
\section{\sysname Design}
We next elaborate on the key components of \sysname.

\subsection{Content-aware Segmented Generation}
\label{sec:con_infer}


We first present our content-aware segmented LLM generation.  The keys are two: (\textit{i}) identifying the right time for segmentation, and (\textit{ii})  context switching, both of which must be done efficiently. 

\begin{figure}[t!]
\centering
\includegraphics[width=0.49\textwidth]{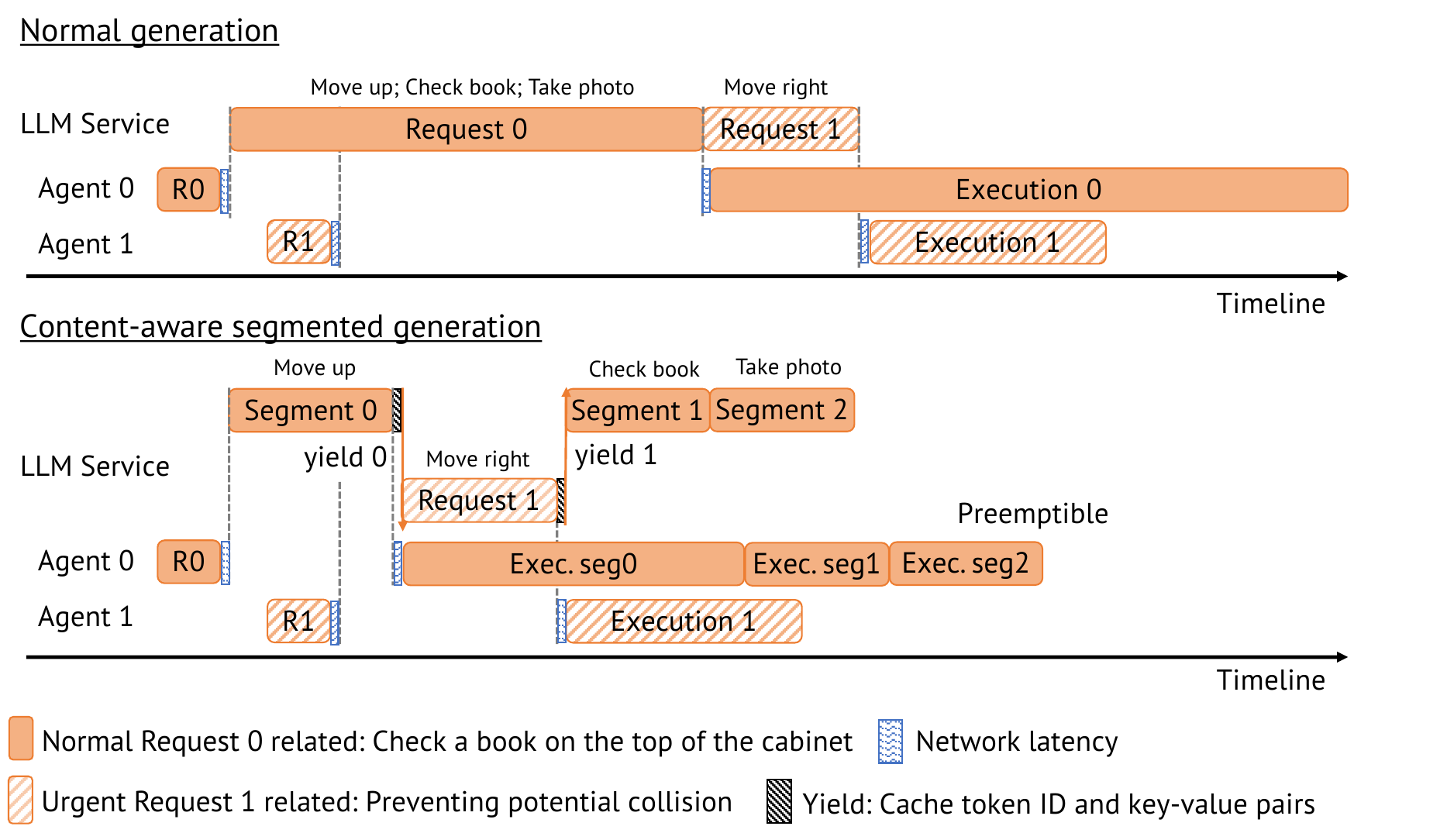}
\caption{\textmd{\small A Case Comparison of Normal Generation vs. Context-aware Segmented Generation: Under normal generation mode, Urgent Request 1 cannot be executed in time due to the blocking of Normal Request 0 generation. Content-aware segmented generation optimizes this by releasing resources upon completing Segment 0 of Request 0, thereby allowing the system to process Urgent Request 1. Once Request 1 is processed, the system resumes and completes the subsequent segment of Request 0, which is finished before the completion of Segment 0 execution. There is no network latency shown before Segment 1 and Segment 2 execution since they are paralleled with the execution of the previous segment.}}
\label{fig:con_infer}
\end{figure}

\myparagraph{Segmentation.}
As shown in \autoref{fig:con_infer}, when \sysname detects some executable skill(s) in the generated content, it suspends the generation to free resources to serve other requests. 
This is based on the insight that there is no urgency to generate the rest of the response because the robot takes a much longer time to execute the skills.

Unlike traditional LLM generation, which produces tokens until an EOS token appears and then detokenizes all the generated tokens, \sysname detokenizes token IDs as they are generated in order to check an executable skill has been generated,  e.g., \skill{pick([obj],\{\})} in \cite{wang2024llm} and \skill{mf[distance:int]} in \cite{chen2023typefly}. 
As clarified in \S\ref{subsec:robot_latency}, the robot skill sets are often provided in the robotic requests, and we leverage that information for executable skill identification. In our implementation, we assume users pre-store the fixed components of requests like robot skill sets, guidance for plan generation, and case examples on the server, which eliminates the need to repeatedly extract skill sets from requests at runtime.
As shown in \autoref{tab:latency}, the detokenization process, which represents only about $0.01\%$ of the total generation time, means that our approach incurs negligible computational overhead.

\myparagraph{Context switching.} When suspending an generation, \sysname retains its KV cache as the intermediate state as well as the tokenized output so far, which constitutes the context of the generation. When resuming the generation, \sysname only needs to restore its context, eliminating the need for prefilling. We chose to use tokenized output as part of the context for two reasons. First, it obviates retokenization at the time of resumption, which reduces the latency, especially for generations of long prompts. 
Second, it prevents mismatches that can occur during retokenization. For instance, encoding the token IDs "[  30,  344,  493, 2239,  \textbf{873,  419}, 2575]" produces the text "?iv('book')==True", but retokenizing this text might yield a different token sequence like "[30,    344,    493,   2239,  \textbf{99991},   2575]". 
The memory overhead for storing the tokenized output is reasonably modest, requiring just 144 bytes to store 13 tokens for a concise robot plan.

Saving the context of a suspended generation adds pressure to the memory hierarchy of \sysname. As active generations are likely to exhaust the rather limited GPU High-bandwidth Memory (HBM), the context of a suspended generation is likely to be evicted by the underlying memory management subsystem from GPU HBM and reside in the more abundant host memory. 
This leads to overhead when the suspended generation is resumed: \sysname must retrieve its context from the host memory to the GPU HBM. 
However, for lengthy robotic requests, storing the KV cache remains more efficient than re-prefilling.
For instance, the context for a single robot control generation using Llama3-8B-float16 includes $170.35\ MB$ of the KV cache and $0.012\ MB$ of the tokenized output. Transferring this data from host memory to GPU HBM takes $9.50\ ms$, less than $3\%$ of the total generation time ($\sim 370\ ms$), whereas re-prefilling the inputs takes significantly longer at $133.31\ ms$.

\subsection{Segmented Scheduling}
\label{sec:seg_schedule}
\sysname features a segmented LLM scheduling mechanism designed to control the generations to meet the \rt requirements of robotic requests. We next introduce the task model for our \sysname. 


We define the set of LLM tasks as the collection of requests received from robotic agents, denoted by $\mathbb{T}$.
Requests can arrive concurrently and continuously.
We represent the $i$th received request as $r^{i}$, where $i$ from 0 to $I$.
With fixed server resources, \sysname aims to maximize the time utility of all received requests. The overall objective can be formulated as follows, where $W(\cdot)$ is the actual response time of one request, and $TUF^{i}(\cdot)$ is the time utility function of each request $r^{i}$.

\begin{equation}
\begin{aligned}
& \text{max} \ \ \ \ \sum_{r^{i} \in \mathbb{T}} TUF^{i}(W(r^{i}))\\
\end{aligned}
\label{eq:overall_goal}
\end{equation}

\begin{figure}[t!]
\centering
\includegraphics[width=0.45\textwidth]{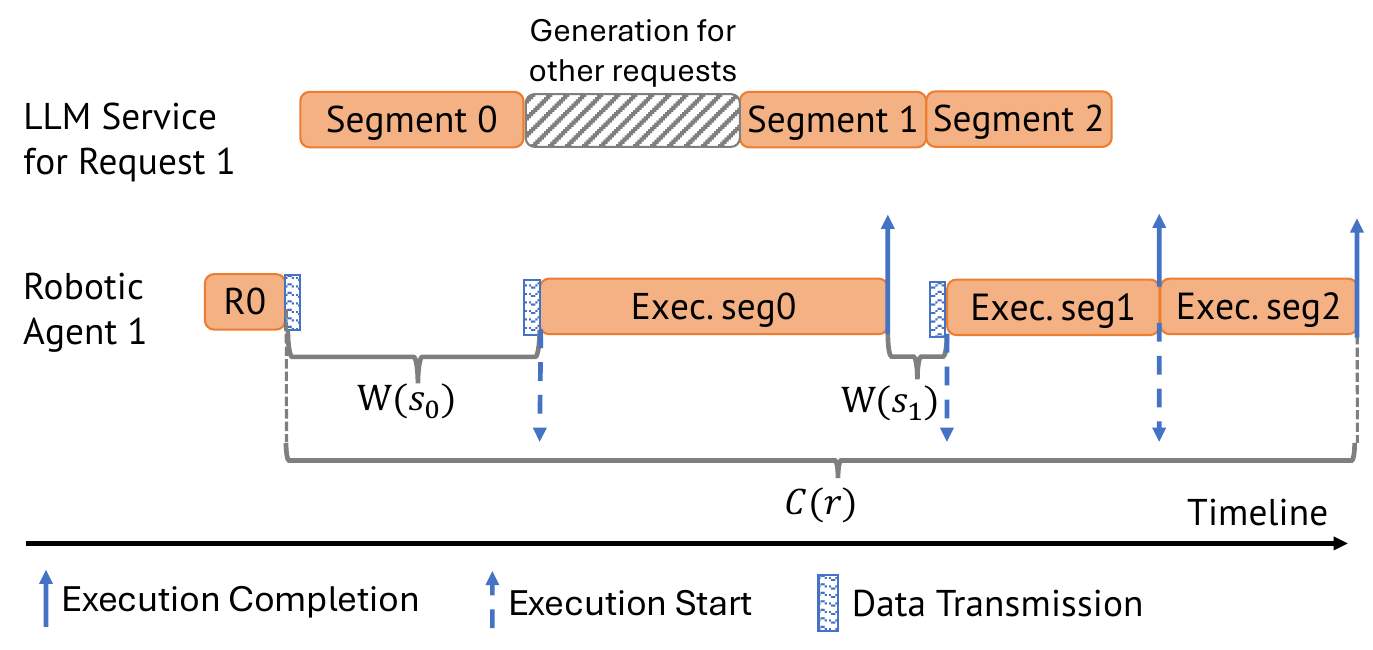}
\caption{\textmd{\small Three user-perceptible latencies for a robotic request: (i) Request response time $W(s_0)$: the time taken to perform the first action, which is also the waiting time of segment 0. (ii) Robot waiting time $\sum_{k=0}^{K}W(s_{k})$, $W(s_0)+ W(s_1)$ in the figure: the cumulative waiting time introduced by LLM planning, represented as the sum of waiting time for all segments. For normal LLM generation, this equals the request response time. (iii) Task completion time  $C(r)$: the total duration encompassing the robot waiting time and execution time.}}
\label{fig:llm_time}
\end{figure}

\sysname generates a response in multiple segments. Hence, we represent the LLM output with a set of sequential segments $[s_{0}, s_{1},...,s_{K}]$, where $K$ is the total number of segments. Given request $r^{i}$,  the $k^{th}$ segment of the LLM output is $s_k^{i}$.
Each segment includes some executable skill(s) and as a result, leads to its own robotic action.
We define the waiting time of a segment as the time between the completion of the preceding segment's action, or the reception of the request in the case of the very first segment, and the start of its own action, as shown in ~\autoref{fig:llm_time}.


Segmented generation may delay subsequent actions and increase the task completion time. The task completion time comprises the cumulative waiting time $W(\cdot)$ and the execution time $E(\cdot)$ of each segment. 
Balancing these objectives—maximizing the time utility of the response time while minimizing task completion time across all requests—poses a significant challenge. To address this, we reformulate the optimization problem to focus on maximizing the time utility of waiting times across all segments, as expressed in the following equation:

\begin{equation}
\max \sum_{r^i \in \mathbb{T}} \big(TUF_0^i (W(s_0^i)) +\sum_{k=1}^{K^{i}} TUF_1^i (W(s_k^i) ) \big )
\label{eq:overall_goal_seg}
\end{equation}



\noindent
\textbf{Theorem 1} \textit{Assuming that the TUFs are monotonically non-increasing, the optimal solution to Eq.~\ref{eq:overall_goal_seg} is Pareto optimal with respect to both the request completion time and the first-segment time utility.}

We prove Theorem 1 in \autoref{appendix:solution_equ_proof}. It is worth noting that in many practical scenarios, TUFs are indeed monotonically non-increasing~\cite{wiki-tuf}, making our solution practicable. 

In our implementation, we use the original TUF provided by the agent as $TUF_0$. For TUF configurations in suspended generations, the parameters $\beta$ and $\alpha$ inherit their values from the original TUF specified by the agent, as the time criticality remains unchanged. To ensure the subsequent response segment is generated before the robot completes the preceding one, we set $ERT$ to zero. Consequently, the $TUF_1$ for each suspended generation is represented as:
$TUF^i_1(t) = \min(\beta^i, \alpha^i \cdot \max(t,0) + \beta^i)$.

\subsection{Execution-based Priority Assignment.}
\label{sec:priority_assign}
Obtaining the optimal solution for the problem formulated in Eq.~\ref{eq:overall_goal_seg} is challenging, primarily due to the unpredictability of future requests and the fact that this problem is NP-hard~\cite{chen1996scheduling}. Given this complexity, we employ a greedy algorithm that selects the generation task with the highest return at each scheduling point, thereby increasing the likelihood of maximizing the aggregate utility. 

However, a significant challenge arises because the segments of each request cannot be determined or predicted in advance due to the probabilistic nature of LLM outputs. This uncertainty makes it difficult to directly apply the above greedy algorithm to our segmented LLM generation. To overcome this, we analyze the time requirements of individual segments and dynamically adjust their priority at runtime based on the content generated in preceding segments.

\myparagraph{PUD-based generation priority.}
Existing literature proposes various utility accrual scheduling algorithms~\cite{wu2004utility, li2006utility, balli2007utility, tidwell2010optimizing} to maximize the aggregate time utility by selecting the task with the highest return as early as possible. A key concept utilized in these approaches is Potential Utility Density (PUD), which quantifies task returns. PUD represents the potential utility per unit of time, defined as the expected utility of running a job normalized by its expected processing duration.
We similarly employ PUD-based priority assignment, adapting it to suit our segmented LLM generation tasks.

Specifically, we establish a general priority assignment equation for both initial and suspended generations based on PUD.
As shown in Eq.~\ref{eq:pri_set}, $t$ indicates the current time, $G(\cdot)$ represents the generation time duration. We estimate the generation time for each segment based on the average generation time calculated from profiling data across diverse requests. $W(s_k^i;t)$ denotes the waiting time for segment $s_k^i$ when its generation begins at time $t$, while $\frac{TUF(\cdot)}{G(\cdot)}$ is the utility accrued per unit of time.
To differentiate tasks with the same initial utility but different slack times, we incorporate slack time $L(\cdot)$ as a scaling factor in the priority calculation, where $L(\cdot)$ represents the remaining time until the expected completion of generation.

\begin{equation}
\begin{aligned}
Pri(s_k^i, t) = \frac{\delta_{k} TUF^i_0(W(s_k^i;t)) + (1 - \delta_{k}) TUF^i_1(W(s_k^i;t))}{G(s_k^i)L(s_k^i)}
\end{aligned}
\label{eq:pri_set}
\end{equation}
\[
\delta_{k} =
\begin{cases} 
1, & \text{if } k = 0, \\
0, & \text{if } k > 0.
\end{cases}
\]

\myparagraph{Choice of priority parameters.}
Determining the waiting time for suspended generation, $W(s_k^i;t)$ with $k>0$, is crucial for priority assignments. It typically depends on the execution completion time of the previous segment. Since priorities must be determined before the preceding segment completes execution, we estimate the completion time dynamically at runtime instead of relying on actual completion time.

The completion time is the sum of the generation completion time and execution time, with only the latter requiring estimation.
The execution time $E(\cdot)$ for each segment (i.e., robotic plan) is estimated using a predictive time model, constructed from offline profiling that records the execution time of each specific robotic action (e.g., move forward, turn left) with different parameters (e.g., moving distance, turning degree).
Execution time for certain actions varies with the scenario. For instance, a search operation is faster if the target object is directly in front of the robot but slower if it’s behind, requiring the robot to turn. For this situation, we estimate the minimum execution time to ensure the plan can be generated in time, even in the worst case.




\myparagraph{Dynamic priority adjustment.}
Our scheduler assigns the initial priority when a request is received and updates the priority function parameters once a new segment is generated. Notably, before fetching one task from the queue for processing, our scheduler will also update the priority of all the tasks in the queue since our priority function Eq.~\ref{eq:pri_set} depends on the current time t. The overhead incurred by priority updates is reasonably modest, as will be shown in \S\ref{sec:eva_overhead}.

\subsection{Latency-guided Batch Size Selection.}
\label{sec:batching}
In this section, we illustrate how our scheduler identifies whether to dispatch the highest-priority generations from the queue to the LLM inference engine. A primary challenge in this process is controlling the batch size within the inference engine, as the impact of batch size on model inference latency is difficult to estimate \cite{zhang2023shepherd, jiang2023coedge}, especially for LLM inference with variable output lengths \cite{jin2023s, zheng2024response}.
Previous work ExeGPT~\cite{oh2024exegpt} determines optimal LLM batch sizes under time constraints via simulator-scheduler interactions but incurs substantial profiling overhead and introduces runtime simulation delays when adjusting batch sizes.

Fortunately, our segmented scheduling approach offers an opportunity to estimate the worst-case scenario of batch influence, as the length bound for each segment is more predictable than the length of the entire LLM output for one request. Specifically, we assess whether the most urgent generation in the execution pool can be completed within its remaining time budget. 
We estimate its worst-case execution time (WCET) by dividing the remaining token length by the generation speed. The remaining token length is calculated by subtracting the current generated token length from the estimated maximum token length for a segment. The generation speed is derived from the time taken to generate the most recent tokens.
Meanwhile, we monitor the free memory resources to ensure that the GPU memory can support the new generation. By introducing a lag for one generation, this method offers a profiling-independent strategy for effectively assessing resource utilization and deciding on the addition of new generations at run-time.

\section{Implementation}
\label{sec:impl}
We next describe our implementation of \sysname on which the evaluation is based. 

\myparagraph{Customized LLM inference engine.}
We build a new LLM inference engine to support our segmented generation approach by customizing a generation stop checker of the open-source vLLM codebase~\cite{vllm}. We select vLLM as our base engine due to its support for continuous batching and its ability to integrate advanced memory optimization techniques, such as PagedAttention~\cite{kwon2023efficient}.
We adapt the stopping rule in the checker to suit various robotic applications. For example, in robot control, we use regular expression matching to detect whether a robot action appears in the current LLM-generated plan.

\myparagraph{Priority queue.}
We adopt a priority queue to store initial and suspended generations. This queue is implemented using \textit{PriorityQueue} and \textit{multiprocessing}, allowing safe data sharing between processes for scheduling. We define a task class to encapsulate the priority and task details for each item in the queue. 
When \sysname receives requests from different agents, each initial generation for the request is tagged with a task ID, agent ID, and arrival time before being added to the priority queue. 

\myparagraph{LLM inference scheduler.}
Our implementation employs a scheduler to queue suspended generations and retrieve them for inference. The scheduler updates the priority of the tasks in the queue using our priority assignment algorithm before fetching the tasks, ensuring the alignment of priorities with the current time (\S\ref{sec:priority_assign}). As shown in \S\ref{sec:eva_overhead}, the computational overhead of these priority updates is minimal, allowing efficient task management.

 \section{Dataset}
\label{sec:dataset}
To the best of our knowledge, no publicly available benchmarks exist for real-world time traces in multi-agent LLM serving systems. To address this evaluation gap, we present \datasetname, a dataset collection system designed to construct time traces for multiple LLM-powered robots, aiding in the evaluation of LLM serving systems. 

Derived from practical evaluation requirements, we establish several key design objectives crucial for tailoring the dataset collection. These include: (\textit{i}) Authenticity: ensuring the traces accurately reflect real-world applications; (\textit{ii}) Flexibility: enabling the creation of diverse workloads, such as orchestrating tasks for 50 robots; (\textit{iii}) Independence: decoupling the utilized model (e.g., Llama 3) from the trace construction.

We adopt several design choices aligned with our goals: (\textit{i}) We collect data from task execution with a real LLM-powered robot application (e.g., Typefly \cite{chen2023typefly}) and a real LLM model (e.g., Llama3 \cite{llama3}). (\textit{ii}) Utilizing this data, we structure the workload in a programmable manner that allows for flexible adjustments. (\textit{iii}) We develop a time model for robot execution, ensuring accurate execution time estimation and simulation, even if utilizing other LLMs to generate the operational plans.



\begin{table}[t]
  \caption{\textmd{\smaller Benchmark tasks used in the dataset collection.}}
  \label{tab:task_list}
  \footnotesize
  \begin{tabular}{|m{1.8cm}|m{1.1cm}|m{4cm}|}
    \hline
    \textbf{Categories} & \textbf{Trace ID} & \textbf{Task Description}\\
     \hline
     \hline
\multirow{5}{2cm}{Normal request for drone}
    & 1 & Look for the cat\\\cline{2-3}
    & 2 & Check for keys or wallets \\\cline{2-3}
    & 3 & Check the book on the cabinet \\\cline{2-3}
    & 4 & Find a toy under the bed \\\cline{2-3}
    & 5 & Find a tool for cutting paper 
\\\hline
\multirow{3}{2cm}{Urgent request for drone}
& 6 & Avoiding human injury \\\cline{2-3}
& 7 & Preventing potential collision \\\cline{2-3}
& 8 & Evading animal pursuit
\\\hline
\multirow{3}{2cm}{Complex request for robot arm}
& 9 &  Object stacking \\\cline{2-3}
& 10 & Desk cleaning \\\cline{2-3}
& 11 & Fruit and vegetable classification
\\\hline
  \end{tabular}
\end{table}


\begin{table}[t]
    \caption{\textmd{\smaller Different Workloads for Evaluation. (WID: Workload ID, EPS: Event Per Second, Max TPE: Maximum Task number Per Event, AN: Agent Number). WID 1 incorporates tasks with varying TUF characteristics. WID 2 differs from the first by offering a higher degree of task parallelism. WID 3 is designed for more complex tasks involving the robot arm.}}
    \label{tab:data_sample}
    \footnotesize
  \begin{tabular}{@{}>{\raggedright\arraybackslash}p{0.5cm}>{\raggedright\arraybackslash}p{2cm}>{\raggedright\arraybackslash}p{0.5cm}>  {\raggedright\arraybackslash}p{0.7cm}>  {\raggedright\arraybackslash}p{1.5cm}>{\raggedright\arraybackslash}p{0.5cm}>{\raggedright\arraybackslash}p{0.7cm}@{}}
\toprule
    \textbf{WID} & \textbf{Categories} & \textbf{EPS} & \textbf{Max TPE} & \textbf{Trace ID} & \textbf{AN} & \textbf{Time (s)} \\ 
     \hline
1 & Mixed TUF & 0.25 & 8 & [1-8] & 25 & 260\\ \hline
2 & Parallel Degree & 0.25 & 16 &  [1-8]  & 42 & 300 \\ \hline
3 & Different robot & 0.1 & 8 &  [9-11]  & 40 & 900 \\ \hline
\end{tabular}
\end{table}


\myparagraph{Robot trace collection.} 
We have expanded the task set designed in an open-source LLM-powered drone system \cite{chen2023typefly}, and specifically tailored additional tasks for robot arms. 
As detailed in \autoref{tab:task_list}, the `normal request for drone' tasks feature relatively simple objectives like searching for a cat or locating keys, aimed at testing basic LLM robot control capabilities. The `urgent request for drone' tasks require a more timely response. The `complex request for robot arm' tasks require a higher degree of LLM reasoning, generating longer plans (around 100 tokens) compared to the 20-token outputs for basic drone tasks. These categories span a diverse range of scenarios to thoroughly cover the functional scope of LLM-powered robot systems under both standard and challenging conditions.
In total, we select 11 tasks from two kinds of robots to exemplify these varied task patterns, ensuring our data collection captures both simple and complex task scenarios. For each task, we use a real LLM-powered robotic application \cite{chen2023typefly} with Llama 3-8B-float16 to generate the robot plan. We opt for this model because it can generate valid control instructions for us to run the evaluation. 

\myparagraph{Programmable trace composition.} We structure the workload for LLM-powered robots in a programmable manner, allowing for dynamic adjustments to accommodate various experimental needs. This flexibility includes modifications to both the arrival time distribution of tasks and the criteria for selecting tasks. Based on the collected time traces from $Trace_1$ to $Trace_{11}$, we could structure workloads in multiple formats, such as \{($Trace_1$,$Trace_2$) , $Trace_8$, $Trace_9$\}. Here, the ($Trace_1$,$Trace_2$) indicates that two robots are simultaneously performing the task denoted by $Trace_1$ and $Trace_2$, which helps in assessing the system's performance with parallel tasks. Developers have the flexibility to adjust the maximum task number per event (Max TPE in \autoref{tab:data_sample}) to assess workloads featuring varying degrees of parallelism.
Additionally, we characterize the event trigger traffic within our dataset using a Poisson distribution, with the event arrival rate (event per second) serving as the Poisson parameter. This probabilistic approach aligns with the MLPerf LLM Inference Benchmark \cite{MLPerfLLAMA2_70B_2024} to emulate real-world operational conditions. 
In our evaluation, we employ three distinct workload configurations, as detailed in \autoref{tab:data_sample}, each identified by Workload IDs (WID).



\begin{figure}[t!]
  \centering
  \begin{subfigure}{0.2\textwidth}
    \centering
    \vtop{\null\hbox{\includegraphics[width=\linewidth]{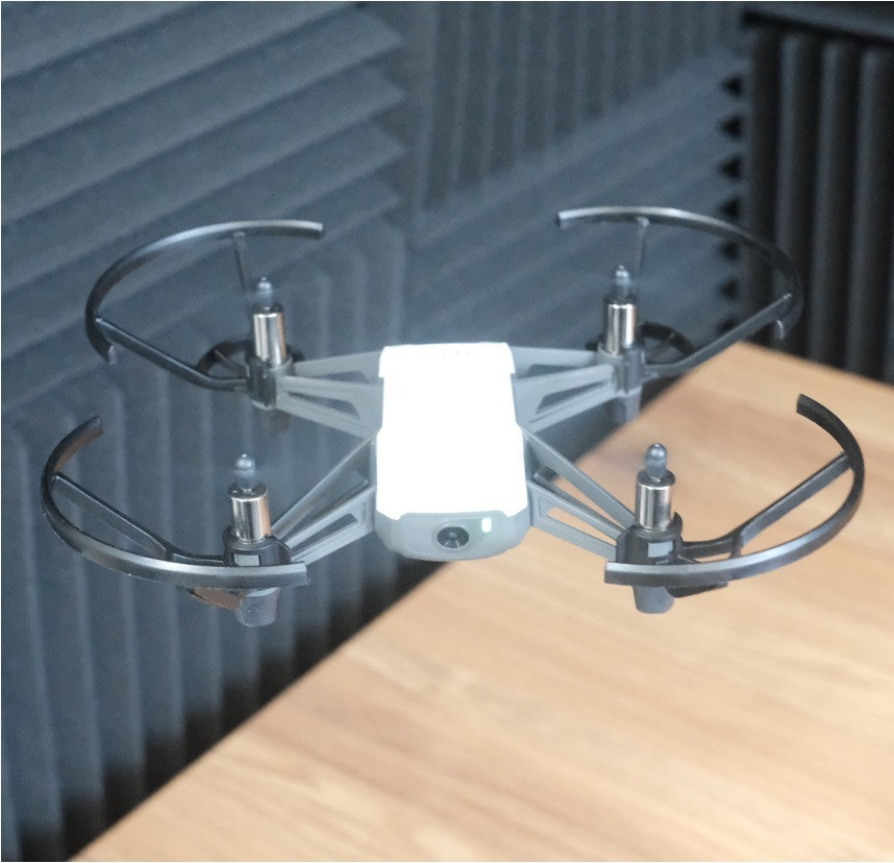}}}
    \caption{\textmd{\small Drone }}
  \end{subfigure}
  \begin{subfigure}{0.14\textwidth}
    \centering
    \vtop{\null\hbox{\includegraphics[width=\linewidth]{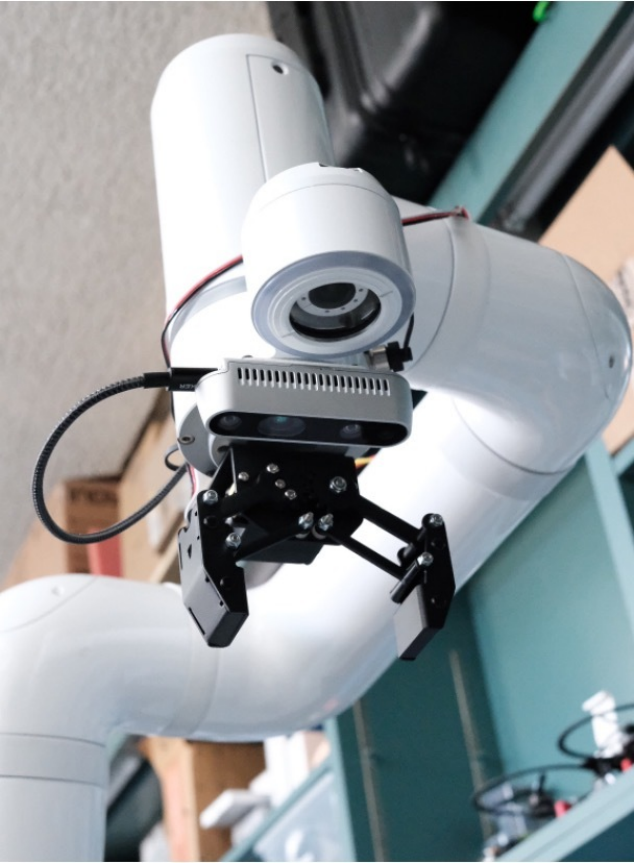}}}
    \caption{\textmd{\small Robot arm}}
  \end{subfigure}
  \caption{\textmd{\small Robots used in data collection: We utilize a Ryze Tech Tello drone \cite{dji2023tellosdk} and a Neuromeka Indy7 Pro robotic arm \cite{robotarmindy}. These platforms are used to profile the actual execution time of various robotic skills. }}
  \label{fig:robot_fig}
\end{figure}
\myparagraph{Execution time model.} 
We then construct a time model to predict the execution duration of each robotic skill and simulate the robotic execution with corresponding delays. We initially profile the actual execution time of each skill on a real drone (i.e., Ryze Tech Tello \cite{dji2023tellosdk}) and a real robot arm (i.e.,  Neuromeka Indy7 Pro \cite{robotarmindy}), as shown in \autoref{fig:robot_fig}, and then develop the time model based on the profiled data.
For simple skills like \skill{move forward}, our model takes the distance or angle as input and provides an estimated execution time. For complex skills like search, we collect execution times across various scenarios, including targets in different positions. During simulation, we randomly sample from this profiled data to reflect real-world variability.
For simpler skills with relatively consistent execution time, we measure the execution time multiple times and use the average as the estimated value.
These time models allow users to quickly simulate robot execution for various plans generated by different LLM models.

\section{Evaluation}
\label{sec:eva}

\begin{figure*}[t!]
  \centering
  \begin{subfigure}{0.33\textwidth}
  \centering
    \includegraphics[width=1\linewidth]{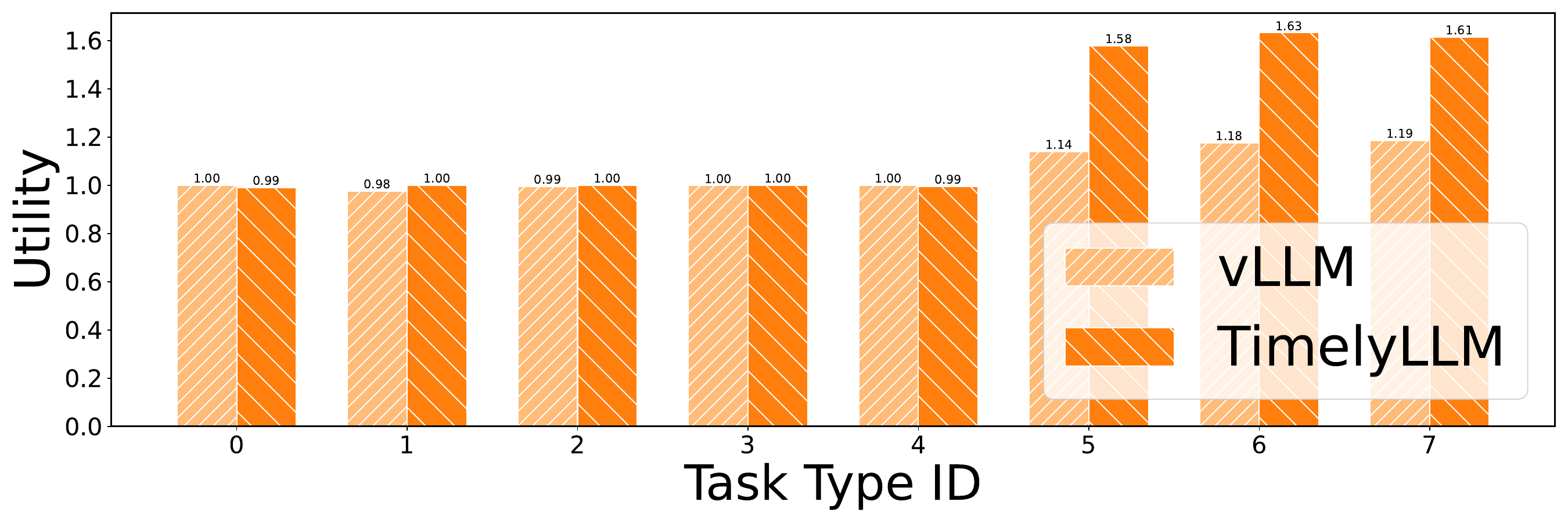}
  \caption{\textmd{\small Time utility-LW.}}
  \end{subfigure}
  \begin{subfigure}{0.33\textwidth}
  \centering
    \includegraphics[width=1\linewidth]{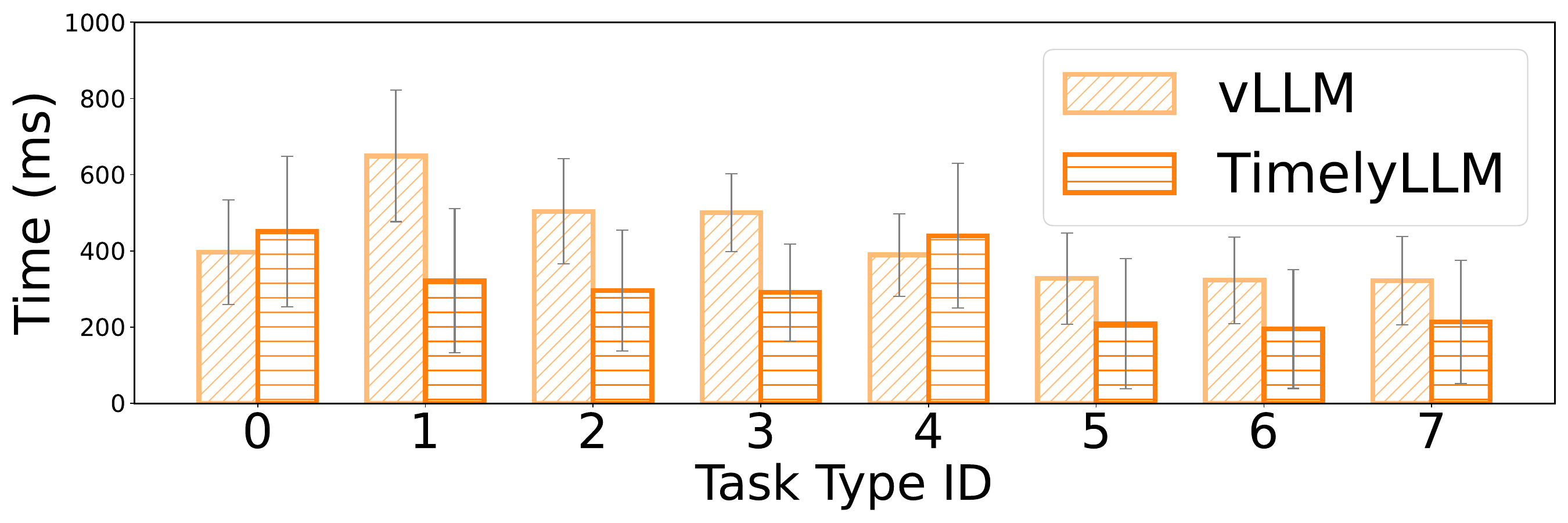}
  \caption{\textmd{\small  Response time-LW.}}
  \end{subfigure}
 \begin{subfigure}{0.33\textwidth}
  \centering
    \includegraphics[width=1\linewidth]{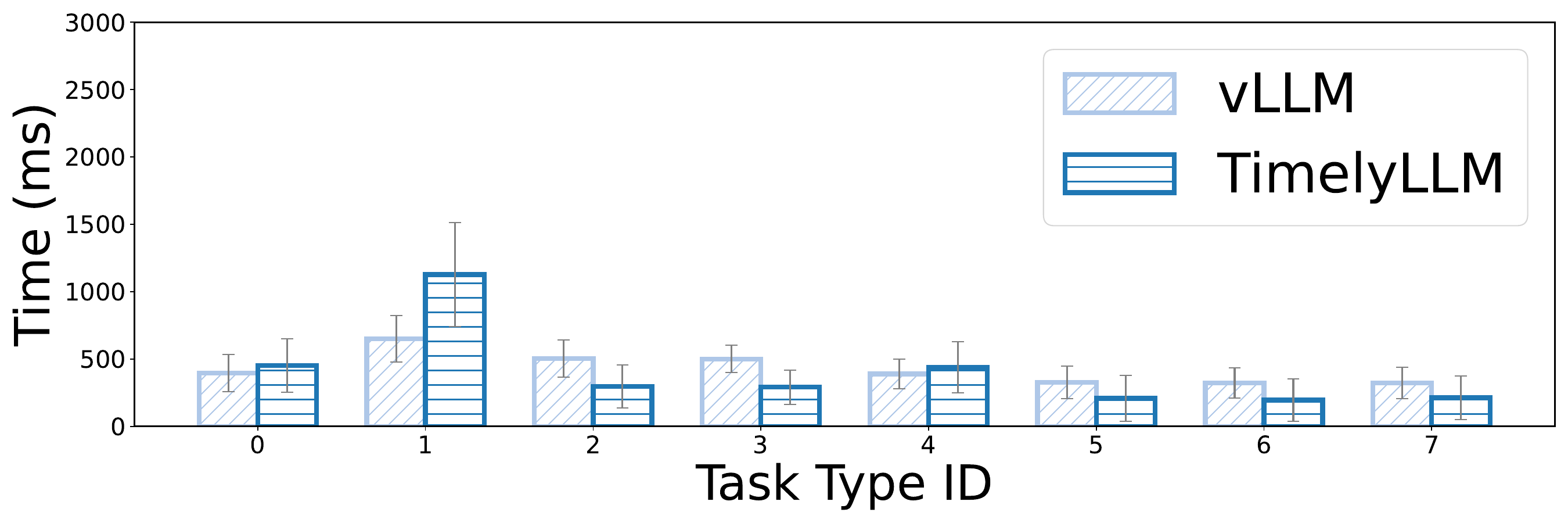}
  \caption{\textmd{\small Waiting time-LW.}}
  \end{subfigure}
  
 \begin{subfigure}{0.33\textwidth}
  \centering
    \includegraphics[width=1\linewidth]{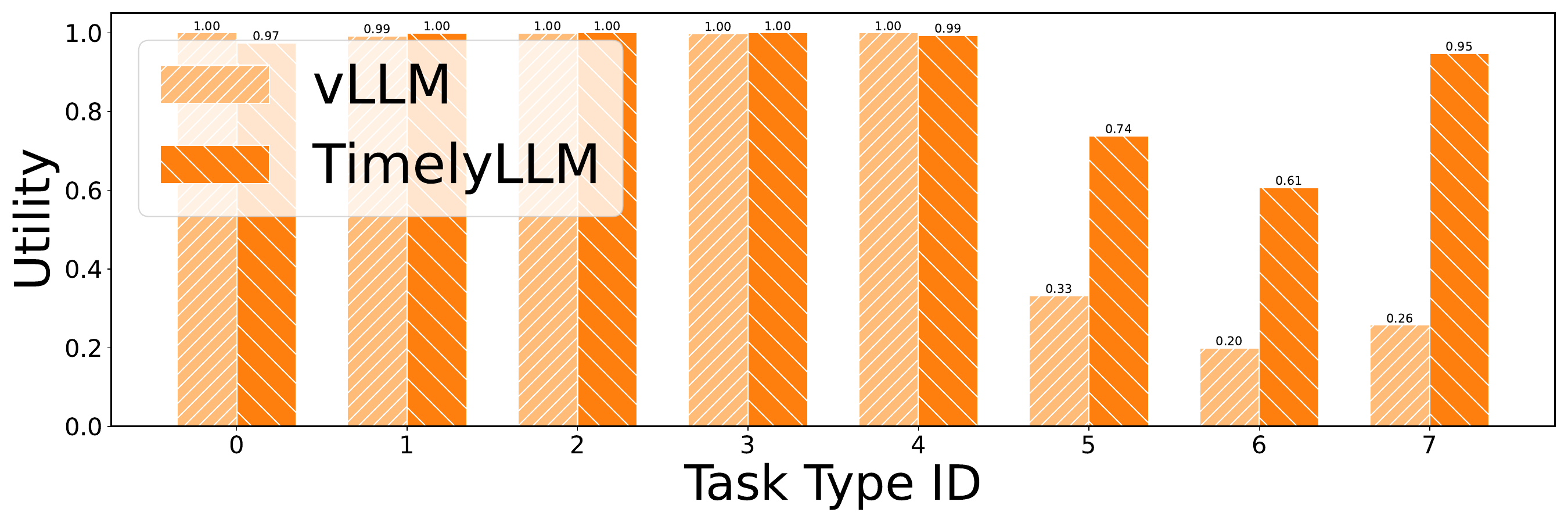}
  \caption{\textmd{\small Time utility-HW.}}
  \end{subfigure}
  \begin{subfigure}{0.33\textwidth}
  \centering
    \includegraphics[width=1\linewidth]{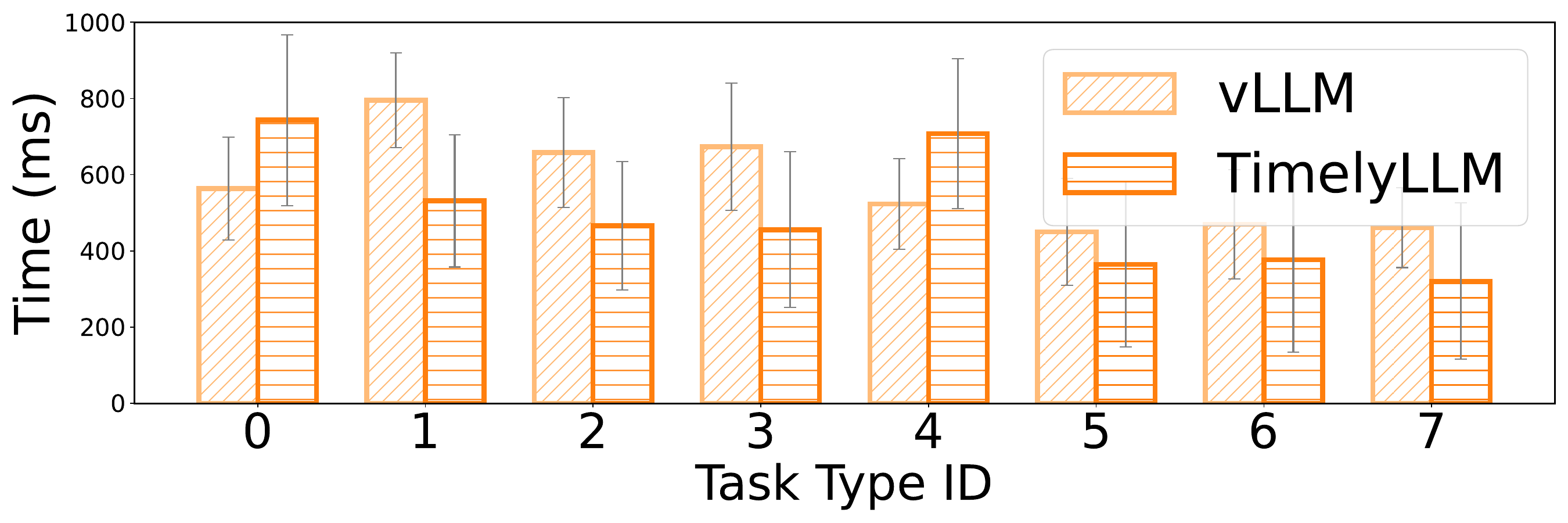}
  \caption{\textmd{\small Response time-HW.}}
  \end{subfigure}
 \begin{subfigure}{0.33\textwidth}
  \centering
    \includegraphics[width=1\linewidth]{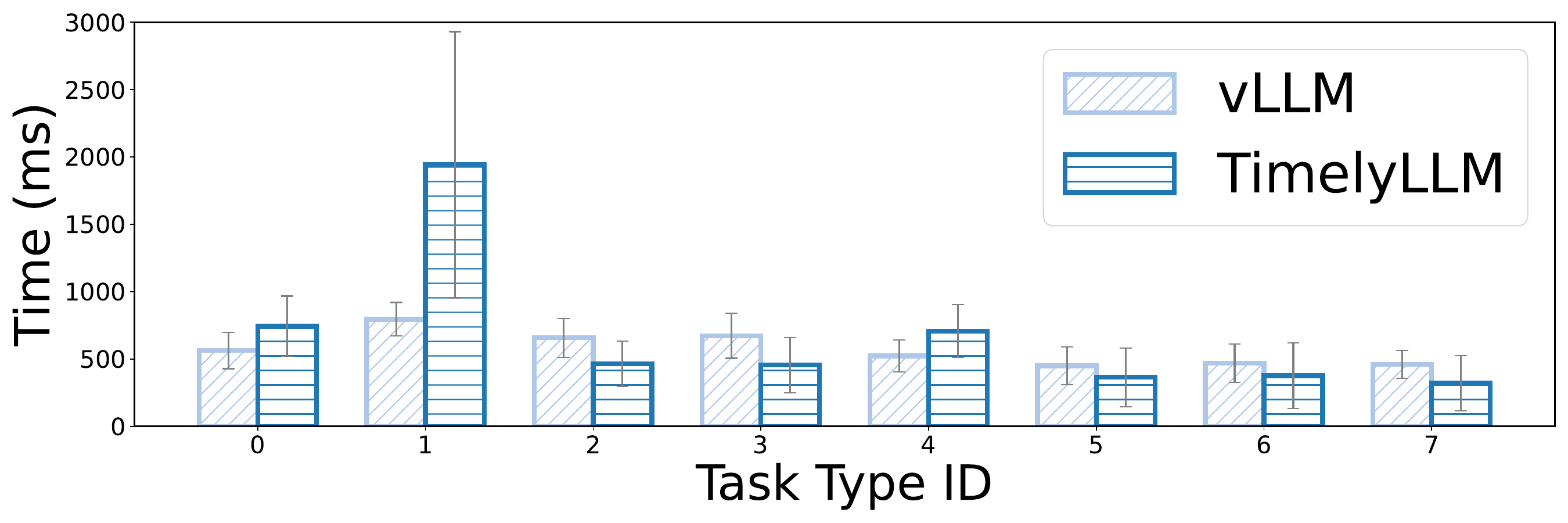}
  \caption{\textmd{\small Waiting time-HW.}}
  \end{subfigure}
  \caption{\textmd{\small End-to-end Performance under different levels of resource contention. (LW: Low Workloads, HW: High Workloads). Based on (a) and (d), \sysname improves the utility significantly over the vLLM baseline on urgent requests. Additionally, the remaining figures demonstrate that \sysname reduces the response time and waiting time across most tasks. }}
  \label{fig:diff-wl}
\end{figure*}



\begin{figure}[t!]
\centering
\includegraphics[width=0.5\textwidth]{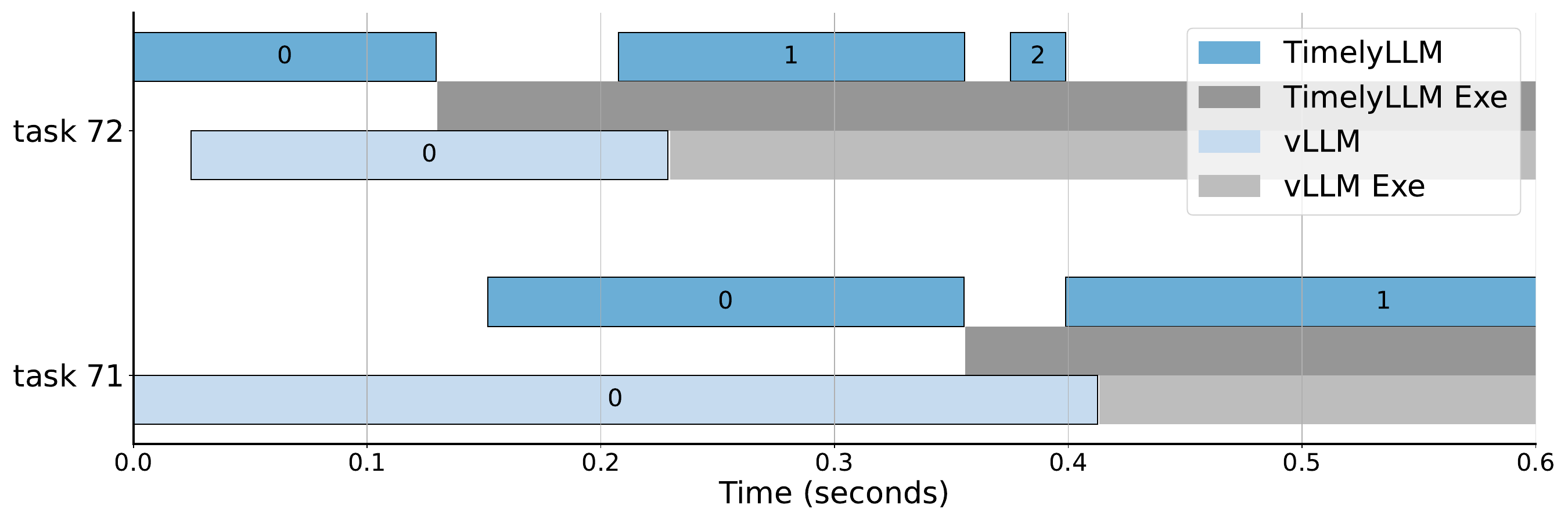}
\caption{\textmd{\small Time trace for Task 71-72 derived from WID 2.}}
\label{fig:time-trace}
\end{figure}

We evaluate \sysname with a set of workloads as described in \S\ref{sec:dataset}. We seek to answer the following questions:
\begin{itemize}[leftmargin=1em]

\item How does \sysname perform in terms of time sensitivity, as measured by the aggregated Time Utility Function (TUF) for each task?

\item How effective of each system component in enhancing \sysname's time sensitivity?

\item How does \sysname perform when applied to more complex planning tasks (i.e., longer response) and other robotic applications with unpredictable segment length?

\item How significant is the time overhead introduced by segmented generation and scheduling?

\end{itemize}

\subsection{Setup}
\label{sec:setup}

\myparagraph{Models.} 
Our experiments employ Llama3-8B-float16, a state-of-the-art open-source LLM series \cite{llama3}, which is capable of generating valid plans for robot control. In our chatbot applications, we also evaluate \sysname on Phi-3-mini-128k~\cite{abdin2024phi}, a model of 3.8 billion parameters. However, we do not evaluate Phi-3 for robot control tasks due to its limited intelligence. We set the temperature in the model inference to 0, ensuring the reproducibility of the results. 

\myparagraph{Hardware.}
We implement our system on an in-house edge server, configured with an NVIDIA RTX 4090 GPU, an AMD Ryzen 9 7950x CPU, and 24 GB host memory. 
We evaluate on a single GPU, but \sysname can still support larger models with multiple GPUs. With multiple GPUs serving a larger model, each portion of the model can store the KV cache on its GPU along with the request ID for fast resumption of the inference.
Since there is no significant data transmission between GPUs during pauses and resumptions, the scheduling performance will not degrade as the number of GPUs increases.

\myparagraph{Baseline and Other Compared Systems}
\begin{itemize}[leftmargin=1em]
    \item vLLM \cite{vllm} (continuous batching): This baseline employs a continuous batching strategy following an FCFS protocol to handle requests from multiple agents. 
    \item vLLM-stream: We enhance vLLM with a streaming execution mechanism specifically designed for robotics applications. Similar to the vLLM, this approach generates the entire task plan continuously but transmits results immediately as soon as an executable action is identified. We employ the same stop rule as \sysname to determine executable actions for transmission.
    \item Other alternative variants of \sysname: We also utilize baselines that are derived from our system but employ different scheduling algorithms such as First-Come, First-Served (FCFS), and Earliest Deadline First (EDF). We will provide a detailed introduction to them in \S\ref{sec:ab_study}.
\end{itemize}

\myparagraph{Metrics.} 
\begin{itemize}[leftmargin=1em]
\item Time utility: This metric represents the time utility value of the LLM's response. We calculate it for each task based on its actual response time (Eq.~\ref{eq:tuf_robot}). It reflects the time sensitivity of the LLM serving system.
 
\item Response time: This metric measures the duration from receiving an agent query to the initial response (e.g., the first action in robot control), quantifying the response efficiency of the LLM serving.

\item Waiting time: This metric represents the total delay introduced by the LLM planning. 
It is defined as the total completion time for a request minus the robot's actual execution time. We adopt this metric instead of task completion time because it directly reflects the serving efficiency of the LLM system, which is the focus of this paper. On the other hand, as robot execution time remains constant for a given LLM-generated plan, improvements in robot waiting time inherently translate to gains in task completion time.

\end{itemize}

\myparagraph{Applications and TUF.} 
We consider two types of robotic applications.
First, \textit{robot control}. Our evaluation primarily focuses on two distinct robotic applications: drones and robot arms. We utilize the dataset delineated in \S\ref{sec:dataset}  and adhere to the TUF parameters as outlined in \S\ref{subsec:app_llm}. For tasks identified with trace IDs [1-5] and [9-11], we adopt the TUF configuration of normal robot tasks, where the $ERT$ is $1s$, $\beta=1$, and $\alpha=-2$. This $\alpha$ value is derived assuming a cut-off time of 1.5 seconds, beyond which the utility becomes negative. Conversely, for tasks with trace IDs [6-8], which are designated as urgent, the TUF configuration is adjusted to an $ERT$ of $200 ms$, $\beta$ at 2, and $\alpha$ at $-6.67$, with the $\alpha$ value computed using a cut-off time of $0.5 s$. We also profile the generation for each segment in the robot control applications, setting the segment generation time to 90 ms for priority assignment and the maximum token length for adaptive batching to 10. 
    
Second, \textit{chatbots}~\cite{meta2024aiassistant} is another robotic application that benefits from LLM. We also evaluate it in our system. 
Notably, our design are not directly applicable to chatbot applications, due to the unpredictable length of generated content in chatbots. 
We support it by generating sentences/paragraph-level segments and prioritizing segments based solely on the estimated reading time).  
For the execution of the chatbot, we mimic the user reading of the generated text by introducing a delay based on the average adult reading speed, estimated at 300 words per minute~\cite{readspeed}. The TUF for a chatbot is the same as robot interaction with humans (i.e., normal robot tasks in \S\ref{subsec:app_llm}), with $ERT=1s$, $\beta=1$  and $\alpha=-2$. 


\subsection{Overall Performance of \sysname}
We start by comparing the end-to-end performance of \sysname with the vLLM baseline. For this evaluation, we measure and record the time utility, actual response time, and the waiting time achieved by both \sysname and the baseline. The recorded results are averaged based on the task type of each request. For the evaluated workloads, we utilize the workloads WID1 and WID2, as described in \S\ref{sec:dataset}. We trigger tasks and mimic the robot execution according to these workloads. We simulate an 8ms network latency by measuring the time each response remains in the execution queue and adding necessary delays. This latency is based on data transmission tests conducted between two devices on a local network.

\autoref{fig:diff-wl} presents various performance metrics under different workloads.
Task types 0-4 are normal requests and Task types 5-8 are urgent requests.
As shown in \autoref{fig:diff-wl}(a), our method achieves the maximum utility for normal tasks, and 81.5\% of the maximum time utility for urgent tasks, whereas the baseline vLLM achieves only 59.5\% of the maximum time utility for urgent tasks. This improvement is attributed to our system's ability to respond to urgent tasks within a shorter duration, as shown in \autoref{fig:diff-wl}(b). 
From \autoref{fig:diff-wl}(c), we observe that \sysname maintains a nearly consistent waiting time with the response time. This shows that for most tasks, our method successfully generates the next segment before the execution of the previous segment, not introducing many waiting periods between consecutive planning segments.
We observe additional LLM planning time in task type 1. This occurs because the planning for Task type 1 only involves short actions like \skill{print}, which can be completed very quickly (around 1 ms), significantly shorter than the time it takes to generate plans. Consequently, this discrepancy introduces increased waiting time between the execution of two consecutive segments. Additionally, segmentation raises the chance of preemption by urgent tasks, further extending the LLM planning time of task type 1.
\autoref{fig:diff-wl}(d) depicts the time utility under conditions of heightened resource contention. Although utility declines due to significant resource contention, our method still improves the average time utility of urgent tasks (5-8) by 191\% relative to the baseline.

We further demonstrate the advantages of \sysname with a time trace derived from the test in WID2, as depicted in \autoref{fig:time-trace}. 
The trace includes the planning and execution of Task 71 and Task 72 using both our method and the baseline approach. Task 71 is a normal request `find a toy', while Task 72 is an urgent task `evading animal pursuit'. Under the FCFS mechanism in vLLM, Task 72 is batched with Task 71 for generation, causing it to exceed its expected response time (ERT) of 200 ms. \sysname prioritizes the generation for Task 72 due to its higher time utility. After generating the initial segment of Task 72, \sysname reallocates resources to Task 71, ensuring it meets its ERT of 1s. The interval between the start of Task 71 and the completion of Task 72's first segment arises due to other urgent tasks (Tasks 73) in the same time duration, which necessitate resource reallocation from Task 71. In summary, our method allows for more fine-grained resource allocation and successfully allocates resources to tasks with higher time utility, thus resulting in better time utility for the whole task set (\autoref{fig:diff-wl}(a)(d)).



\begin{figure}[t!]
\centering
\begin{minipage}{0.24\textwidth}
    \centering
    \includegraphics[width=1\linewidth]{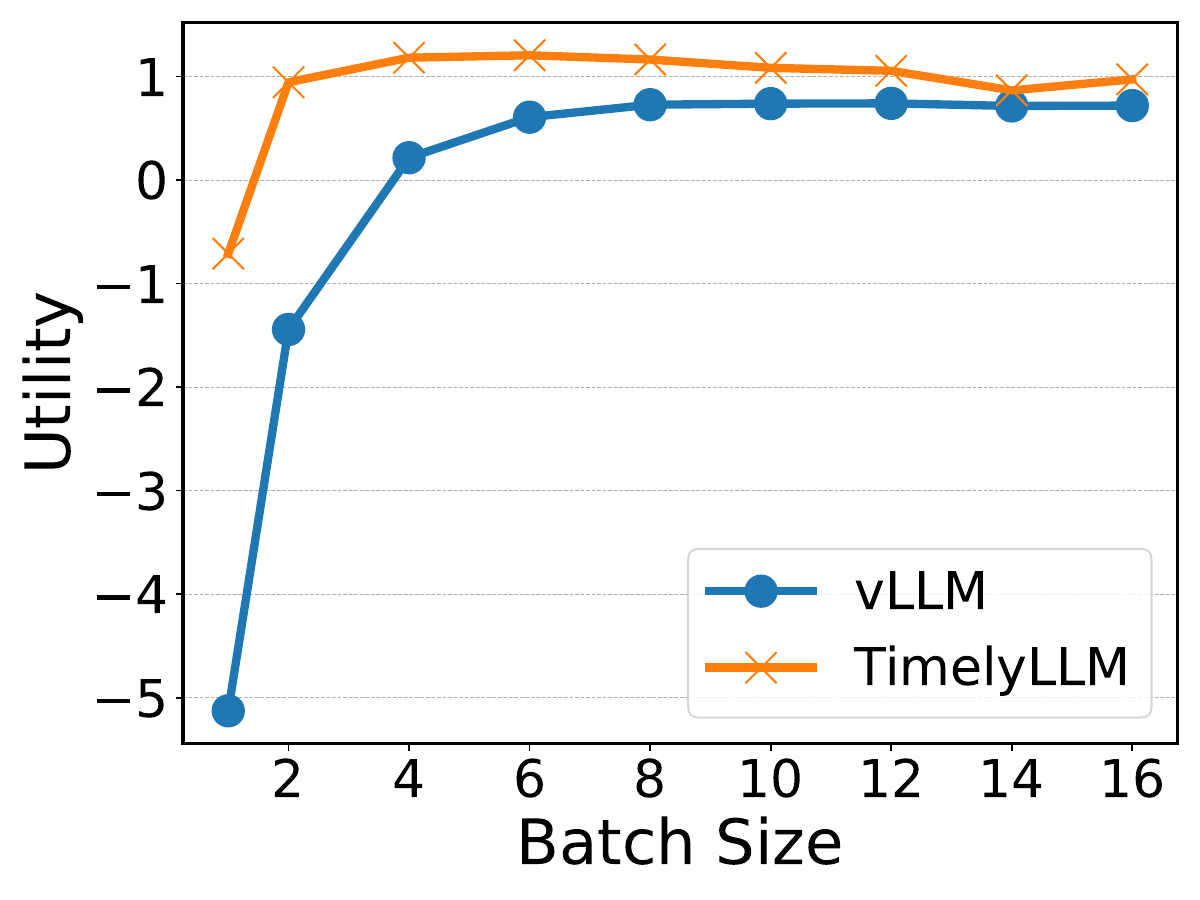}
    \caption{\textmd{\small Performance of \sysname with different levels 
 of throughput (batch size)}}
    \label{fig:thr-util}
\end{minipage}
\begin{minipage}{0.23\textwidth}
    \centering
    \includegraphics[width=1\linewidth]{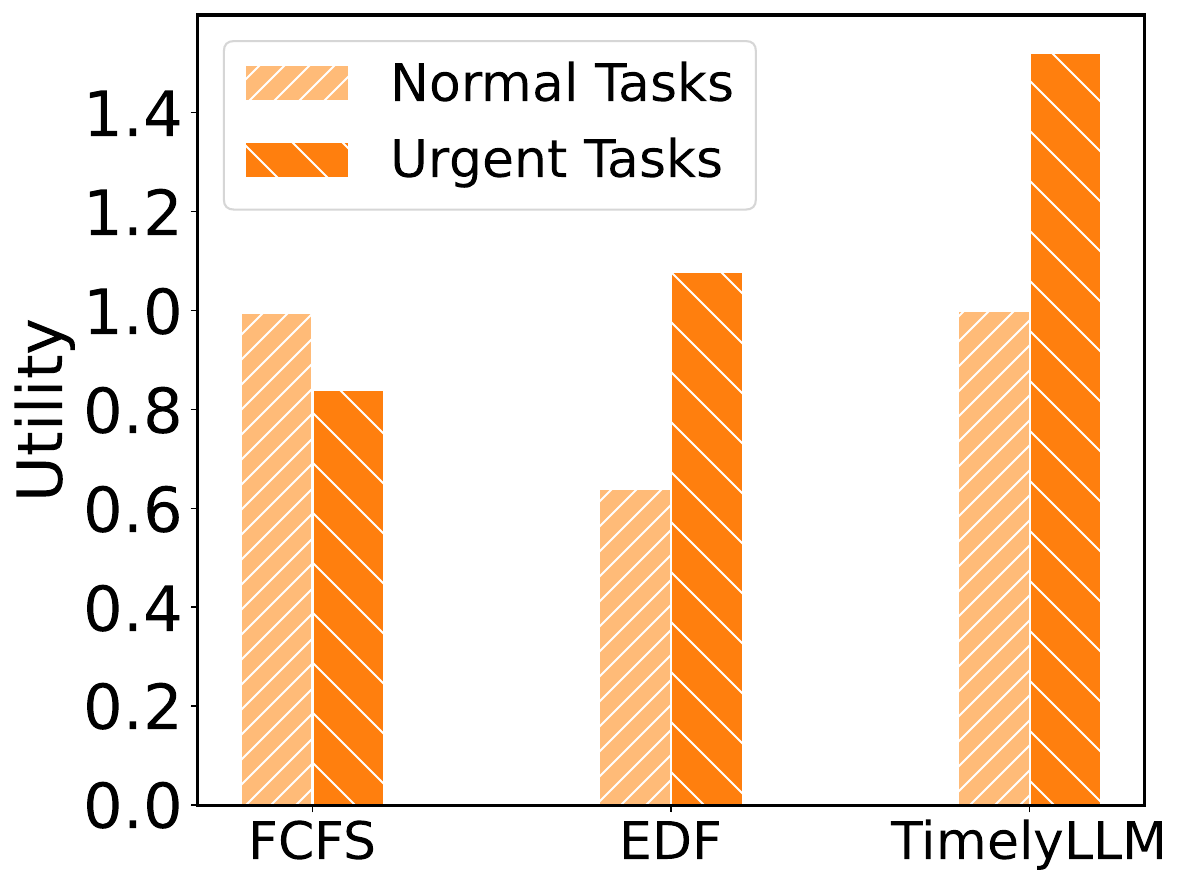}
    \caption{\textmd{\small Performance of \sysname with different scheduling policies, i.e., FCFS, EDF.}}
    \label{fig:ab_schedule}
\end{minipage}%
\hfill
\end{figure}

\subsection{\sysname Deep Dive}
\label{sec:ab_study}

\subsubsection{Performance with different throughput}
We now evaluate the performance of \sysname across various throughput levels by fixing the maximum batch size rather than employing our adaptive batching method, while the other components of \sysname remain unchanged. We similarly adjust the baseline batch size, capping the number of added requests below a fixed maximum.
To assess a broad spectrum of batch sizes, we select WID2 for our evaluation, which has up to 16 tasks per event. As shown in \autoref{fig:thr-util}, we demonstrate the time utility achieved by both our method and the baseline across these varying batch sizes. It shows that the utility values for both models initially increase sharply with the batch size. Upon reaching a batch size of 6, \sysname achieves its highest utility at 1.21, whereas vLLM reaches its peak utility of 0.74 at a batch size of 12. Beyond these points, there is a slight decline in utility for both models. This is because interference between tasks grows with batch size. When batch size exceeds GPU memory limitations, further increases may not increase the actual number of running generations. Overall, \sysname consistently outperforms vLLM across all tested batch sizes.
\subsubsection{Performance with different scheduling algorithms}
We evaluate the performance of our scheduling algorithm by comparing it to segmented generation with FCFS and EDF algorithms. Each baseline employs our segmented generation approach, maintaining the initial generation priority for suspended generations. From \autoref{fig:ab_schedule}, we observe that EDF prioritizes urgent requests, but neglects normal tasks due to its preference for shorter deadlines. Conversely, FCFS maximizes the utility of normal tasks but performs poorly with urgent tasks. In contrast, \sysname achieves better time utility for urgent tasks ($183\%$ relative to FCFS and $142\%$ relative to EDF) without compromising the utility of normal tasks, thanks to its efficient prioritization approach based on the scaled utility density.


\subsection{Scalability of \sysname}
\label{sec:eva_scale}

\begin{figure}[t!]
  \centering
  \begin{subfigure}{0.23\textwidth}
    \centering
    \includegraphics[width=\linewidth]{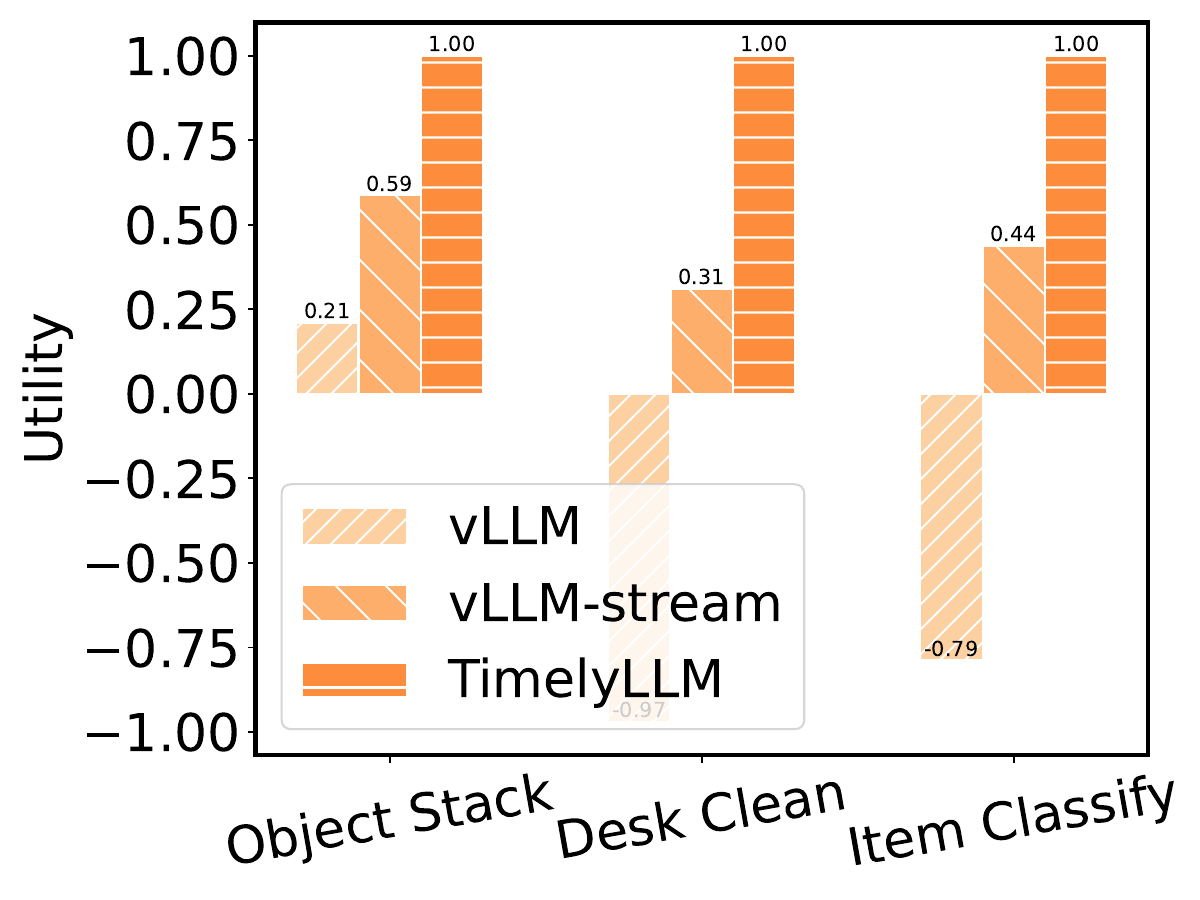}
    \caption{\textmd{\small Time Utility}}
  \end{subfigure}
  \begin{subfigure}{0.23\textwidth}
    \centering
    \includegraphics[width=\linewidth]{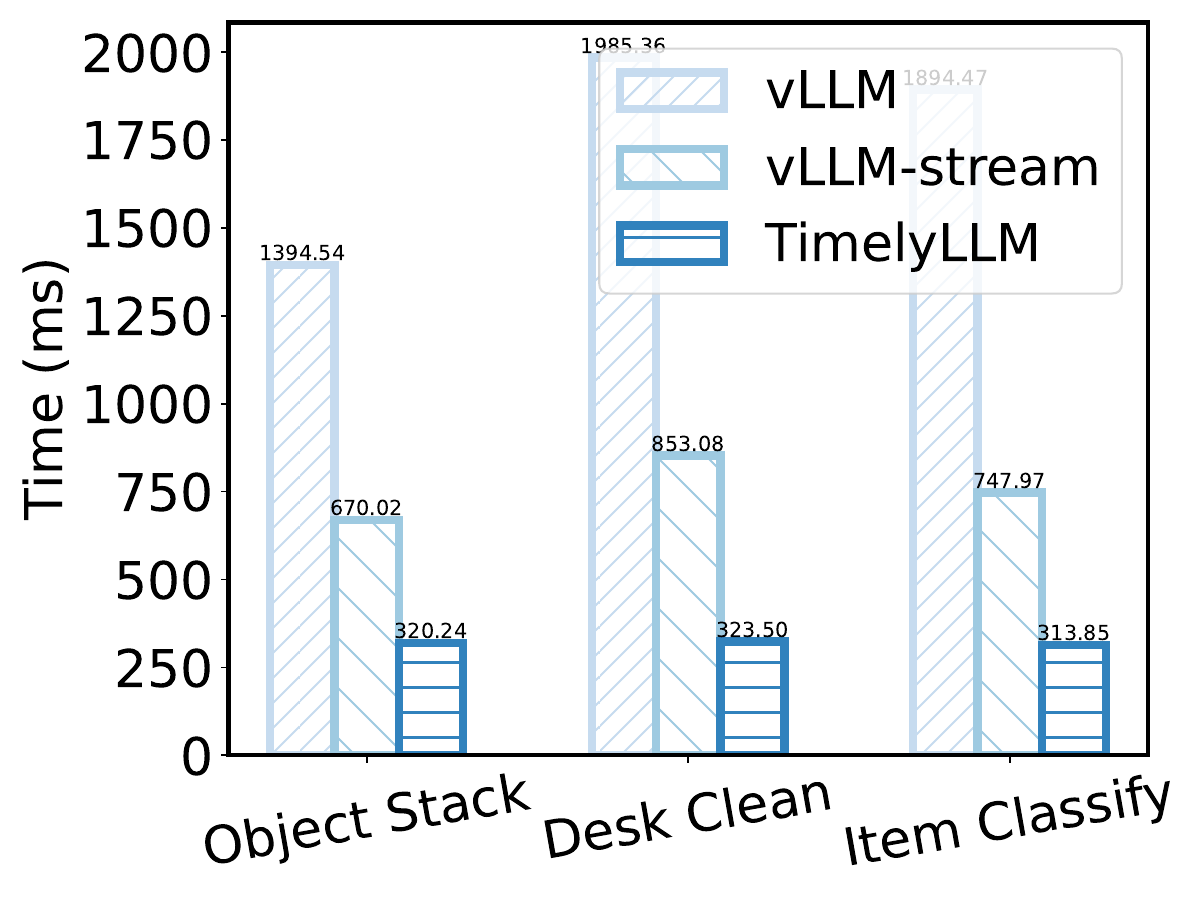}
    \caption{\textmd{\small Waiting time}}
  \end{subfigure}
  \caption{\textmd{\small Performance on robot arm with more complex tasks and longer generated responses.}}
  \label{fig:robot-arm}
\end{figure}

\begin{figure}[t!]
  \centering
  \begin{subfigure}{0.23\textwidth}
    \centering
    \includegraphics[width=\linewidth]{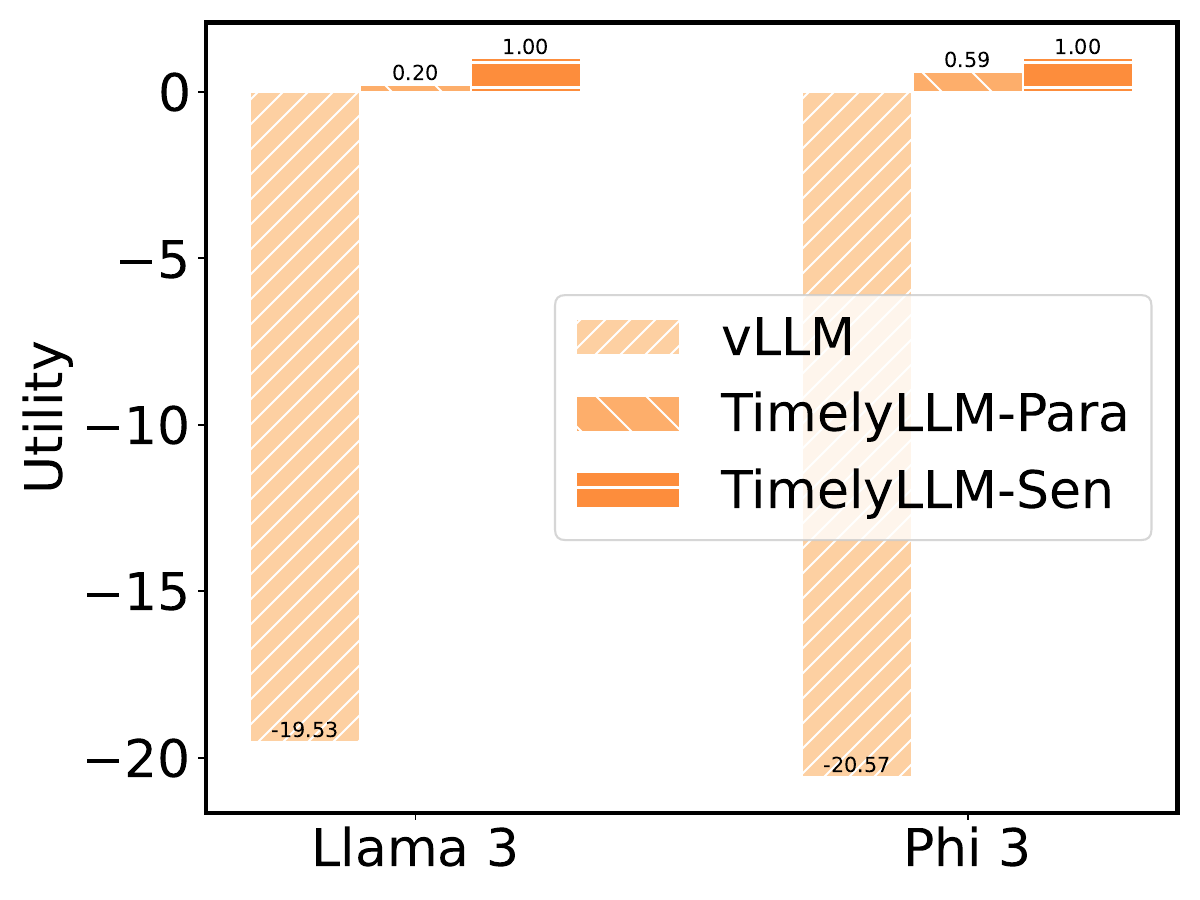}
    \caption{\textmd{\small Time Utility}}
  \end{subfigure}
  \begin{subfigure}{0.23\textwidth}
    \centering
    \includegraphics[width=\linewidth]{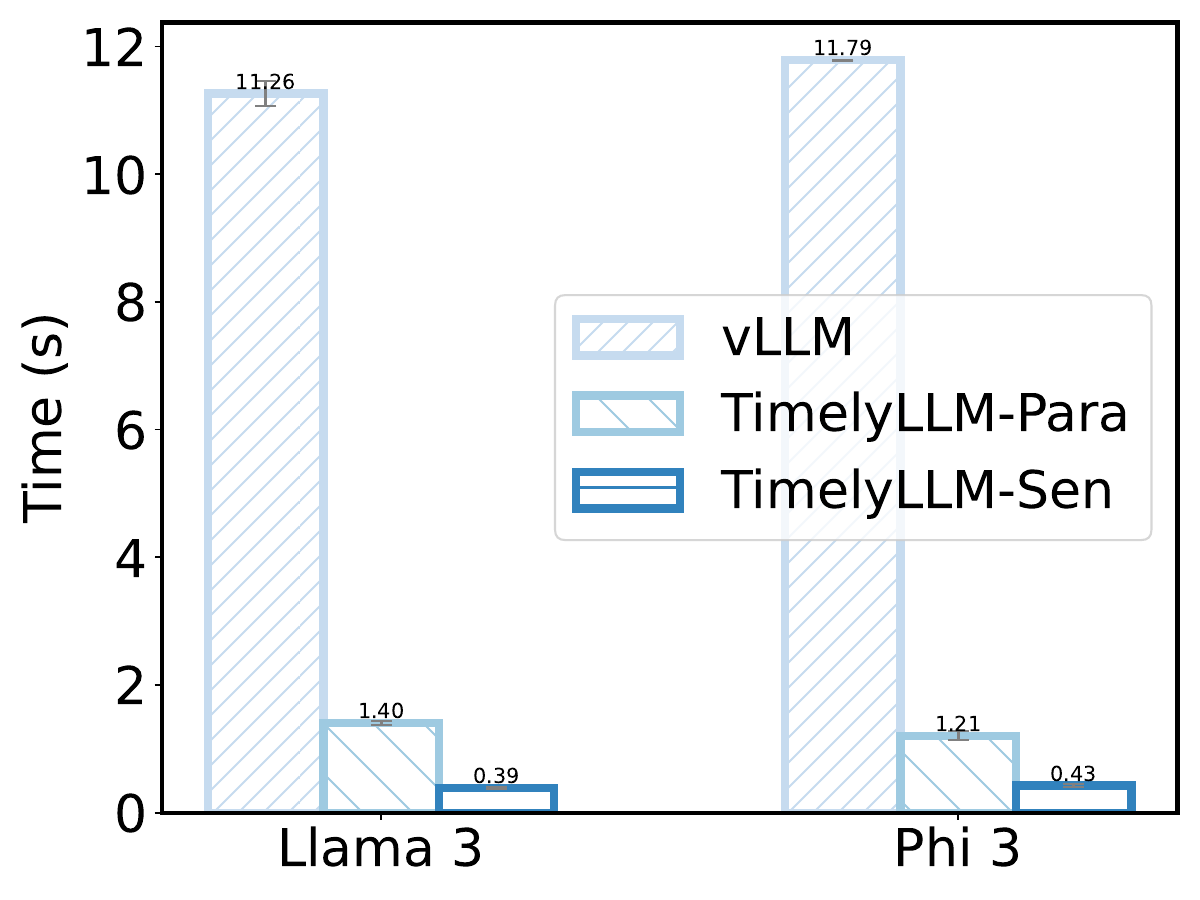}
    \caption{\textmd{\small Waiting time}}
  \end{subfigure}
  \caption{\textmd{\small Performance for chatbot application with Llama3 and Phi3 models.}}
  \label{fig:chatbot}
\end{figure}

\subsubsection{Robot arm task}
\autoref{fig:robot-arm} shows the evaluation results for workload WID 3 with robot arm tasks. The primary distinction between robot arm and drone tasks is the complexity and sequence length of the generated responses. For example, the response for the box stacking task is a long sequence where the robot checks for and manipulates red, blue, and green boxes sequentially. 
From \autoref{fig:robot-arm}, we observe that \sysname substantially enhances the time utility of robot arm tasks. For the best case (`Desk Clean' task), \sysname increases utility by $1.97$ and reduces waiting time by $84\%$ compared to vLLM. This is because the longer response/plan provides more segmentation opportunities, thereby gaining more benefits. We also observe that vLLM-stream performs better than vLLM but results in inferior performance compared to \sysname. 
This is because, while vLLM-stream supports the immediate execution of generated plans, it generates the whole plan at once and does not free up resources for other urgent requests. In contrast, \sysname effectively halts tasks that have already partial plans for execution, reallocating resources to urgent tasks.

\subsubsection{Chatbot task}
\label{sec:eva_chatbot}
We evaluate \sysname's performance on chatbot applications by using a story generation dataset, WritingPrompt \cite{fan2018hierarchical}.
\autoref{fig:chatbot} shows the time utility and planning delays achieved by \sysname with different segmentation rules on different LLM models. \sysname-Para indicates that generation halts after completing a paragraph, while \sysname-Sen stops generation after finishing a sentence. In the best case (\sysname-Sen on Phi 3), our system achieves up to $21.57$ more utility value and $96\%$ less delay compared to the baseline (vLLM).
We observe that \sysname-Sen achieves the highest time utility and the shortest delay on both Llama3 and Phi3, indicating a quicker response time and less waiting time when the system stops generating after a sentence, compared to stopping after a paragraph (i.e., \sysname-Para) or no segment generation (i.e., vLLM). This is because stopping generation after a sentence allows for a shorter response time with smaller segments of text, which also facilitates more fine-grained scheduling.

We also note that \sysname consistently exhibits high-performance benefits across both LLM models. This performance can be attributed to \sysname's non-intrusive approach, which only involves pausing and resuming the inference procedure, thereby not affecting the underlying procedures of LLM inference mechanisms and the generated content. Intrinsically, \sysname supports a variety of LLMs based on transformer decoder architectures.


\subsection{System Overhead}
\label{sec:eva_overhead}
The system overhead of \sysname is mainly caused by the segmented generation and scheduling. We show both kinds of overheads as follows.
\subsubsection{Segmented generation overhead}
We assess the overhead of segmented generation by conducting separate tests for normal and segmented generation. The overhead specific to segmented generation is determined by subtracting the time taken for normal generation from that of segmented generation. This overhead includes stop-checking and caching/retrieval for token IDs and KV pairs. As shown in Table~\ref{tab:seg_overhead}, the segmented overhead is different among tasks due to the different response lengths and content, but all of them remain below $6\%$ of the total segmented generation time.
\begin{table}[t]
\centering
\caption{\textmd{\small The segmented overhead across eight traces, presented in milliseconds and as a percentage.}}
  \label{tab:seg_overhead}
  \footnotesize
\begin{tabular}{@{}ccccccccc@{}}
\toprule
\textbf{Trace ID} & \textbf{1} & \textbf{2} & \textbf{3} & \textbf{4} & \textbf{5} & \textbf{6} & \textbf{7} & \textbf{8}\\ \midrule
Overhead (ms) & 13.7 & 24.1 & 19.9 & 19.9 & 16.0 & 13.1 & 14.4 & 14.8 \\
Overhead (\%) & 3.0 & 2.6 & 5.6 & 2.8 & 3.2 & 3.7 & 4.1 & 4.2 \\ \bottomrule
\end{tabular}
\end{table}

\subsubsection{Scheduling overhead}
With our scheduling algorithm, we need to calculate the TUF for each task when updating the priority. This happens each time we fetch a task from the queue. We quantify the scheduling overhead by measuring the duration required to update the priority and fetch the task.
From Table\ref{tab:sched_overhead}, we observe an increase in scheduling overhead with the number of tasks. However, it remains minimal, with only a 2ms cost for eight tasks.
\begin{table}[t]
\centering
\caption{\textmd{\small The scheduling overhead for various sizes of task queues.}}
  \label{tab:sched_overhead}
  \footnotesize
\begin{tabular}{@{}cccccc@{}}
\toprule
\textbf{Task Number} & \textbf{1}  & \textbf{2} & \textbf{4} & \textbf{6} & \textbf{8} \\ \midrule
Scheduling Overhead (ms) & 0.58 & 0.83 & 1.25 & 1.58 & 2.03 \\ \bottomrule
\end{tabular}
\end{table}

\section{Related work}

\myparagraph{LLM Request Scheduling.}
State-of-the-art LLM serving systems~\cite{vllm, tensorrt_llm, yu2022orca, agrawal2024taming} implement the FCFS scheduling policy, which does not consider the time-sensitivity of LLM requests.
Recent advancements~\cite{wu2023fast, patke2024one, oh2024exegpt,qiu2024efficient, jin2023s} aim to refine LLM scheduling strategies to provide better LLM services.
For instance, FastServe~\cite{wu2023fast} employs a multi-level feedback queue scheduler that prioritizes jobs based on their cumulative generated length. S3~\cite{jin2023s} optimizes scheduling by predicting the output sequence length. Another study~\cite{fu2024efficient} schedules requests by estimating their predicted relative ranks of output lengths.
Despite these innovations, existing frameworks primarily concentrate on scheduling to optimize overall metrics such as job completion time, throughput, and resource utilization, but often overlook the time requirements of each request. Our study bridges this gap by scheduling LLM generation based on actual time requirements of robotic applications, thereby aligning LLM serving with practical operational needs.

Token-level (i.e., iteration-level) scheduling in LLM serving provides more flexibility in managing the generation process than request-level scheduling, but also may lead to more frequent preemption and thereby higher memory managing costs~\cite{yu2022orca, wu2023fast, shahout2024don}. Recent work~\cite{shahout2024don} mitigates this issue by limiting preemption to the first few iterations.
In our work, we introduce segmented generation and scheduling based on the semantics of LLM-generated response, which supports flexible scheduling but also avoids frequent preemption with significant context switch overhead.

\myparagraph{LLM with External Executors.}
LLMs are increasingly integrated with external executors, encompassing both software (e.g., auxiliary AI models~\cite{shen2024hugginggpt}) and hardware (e.g., robotic systems~\cite{chen2023typefly}), to tackle complex tasks. LLMs with API or function call capabilities are widely employed to enable mobile task automation~\cite{wen2024autodroid, lee2024mobilegpt} and solve complicated AI tasks~\cite{shen2024hugginggpt}. Our contribution to scheduling supports both non-blocking and blocking function calls, where each function call represents an executable skill in our system. 
To handle blocking function calls, which typically depend on return values to guide subsequent generation~\cite{abhyankarinfercept}, our system can support it by utilizing a waiting queue. Our scheduler could temporarily place the blocked requests in the waiting queue, and reinsert it into the priority queue with updated priority when the function is completed.
Notably, our system does not change the generation process of LLM but optimizes its generation and execution mechanism, which will not compromise the accuracy or performance of the LLM tasks.

Another line of research enables the asynchronous execution of plans generated by LLMs~\cite{chen2023typefly,kim2023llm}. In these approaches, the LLM generates continuous responses for a single task without interruption, streaming the outputs directly to executors. While this method is effective in environments with unlimited computational resources, it struggles in scenarios where multiple requests contend for resources on a single LLM server. Such resource contention leads to increased delays in both response latency and overall task completion time. We include a comparison with this kind of approach in our evaluation (vLLM-stream), where our method demonstrates superior performance.

\section{Conclusion}
In this paper, we present \sysname,  a \rt LLM serving system designed for multiple robotic agents. \sysname enables the segmented generation of responses and dynamically schedules each request in accordance with time-sensitive requirements and estimated execution time of robotic tasks. We implement \sysname on an open-source LLM serving framework and conduct evaluations across various robotic applications involving LLM-powered drones and robot arms. 
Our findings demonstrate that \sysname outperforms existing state-of-the-art solutions for LLM serving, delivering superior time utility and reduced waiting time when handling real-world robotic queries.

\section*{Acknowledgments}
This work is supported in part by NSF Athena AI Institute
(Award \#2112562) and Yale University. The authors thank Professor Anurag Khandelwal for his input regarding the mathematical formulation in \S\ref{sec:seg_schedule}.

\newpage
\bibliographystyle{plain}
\bibliography{references}

\newpage
\appendix
\section*{Appendix}
\section{Proof of problem equivalence}
\label{appendix:solution_equ_proof}
We reformulate Eq.~\ref{eq:overall_goal} a constraint to minimize task completion time as follows, where the response time of a request $W(r^{i})$ is represented by the waiting time of its first segment, $W(s^{i}_0)$.
Now, the objective of the formulated problem is to maximize the time utility of the response time for all requests while minimizing the task completion time for each individual request.
\begin{equation}
\begin{aligned}
& \text{max} \ \ \sum_{r^{i} \in \mathbb{T}} TUF^{i}(W(s_{0}^{i})) \ \ \ \ \ \ \ \  \\
&  \ \ \ \text{min}  \ \ \sum_{r^{i}\in \mathbb{T}}C(r^{i})\\
& C(r^{i}) = \sum_{k=0}^{K^{i}} (W(s_{k}^{i}) + E(s_{k}^{i}))\\
\end{aligned}
\label{eq:overall_goal_1}
\end{equation}

\noindent
\textbf{Theorem 1} \textit{Assuming that the TUFs are monotonically non-increasing, the optimal solution to Eq.~\ref{eq:overall_goal_seg} is Pareto optimal with respect to both the request completion time and the first-segment time utility.}

\noindent 
\textbf{Lemma 1. }
For any request $r^{i}$ and segment $s_{k}^{i}$, $TUF^i(W(s_k^i))$ is monotonically non-increasing with respect to $W(s_k^i)$. Formally, if $W(s_k^i) \leq W'(s_k^i)$, then:
\[
TUF^i(W(s_k^i)) \geq TUF^i(W'(s_k^i))
\]

\noindent 
\textbf{Lemma 2. }
Under the optimal scheduling solution to Eq.~\ref{eq:overall_goal_seg}, no request completion time can be further reduced. 

\noindent 
\textbf{Proof. }
Let $x^*$ be the optimal scheduling solution to Eq.~\ref{eq:overall_goal_seg}. 
\[
x^* = arg \operatorname*{max}_{x}  \sum_{r^i \in \mathbb{T}} \big(TUF_0^i (W(s_0^i)) +\sum_{k=1}^{K^{i}} TUF_1^i (W(s_k^i) ) \big )
\]
We denote the objective value achieved by the scheduling solution x as v(x),
\[
v(x) = \sum_{r^i \in \mathbb{T}} \big(TUF_0^i (W(s_0^i;x)) +\sum_{k=1}^{K^{i}} TUF_1^i (W(s_k^i;x) ) \big )
\]
Suppose there exists another solution $x'$ that improves the completion time of at least one request $r^{j}$ without impacting the total time utility of the first segments:
\[
C(r^j;x')< C(r^j;x^*),
\]
Since $C(r^{j}) = \sum_{k=0}^{K^{j}} (W(s_{k}^{j}) + E(s_{k}^{j}))$ and execution time $\sum_{k=0}^{K^{j}} E(s_{k}^{j})$ is a constant, any reduction in $C(r^{j})$ implies a reduction in waiting time for at least one segment $s_k^j$(k>0):
\[
W(s_{k}^{j};x') < W(s_{k}^{j};x^*)
\]
Since no other requests need to be negatively impacted by this change, we have:
\[
TUF_1^j(W(s_k^j;x'))\geq TUF_1^j(W(s_k^j;x^*)
\]
This inequality should be strict if there is a strict waiting time reduction: 
\[
TUF_1^j(W(s_k^j;x')) > TUF_1^j(W(s_k^j;x^*)
\]
No, we have:
\[
v(x') > v(x^*)
\]

\noindent
If $x^*$ is truly optimal for Eq.~\ref{eq:overall_goal_seg}, there could be no feasible solution $x'$ that yields a strictly higher objective value. But we have just shown that by reducing one request’s completion time, we can increase the overall TUF of all segments' waiting time. This contradicts the assumption that $x^*$ is optimal. Hence, we prove that under the optimal solution $x^*$, no request completion time can be further reduced. \\

\noindent 
\textbf{Lemma 3. }
Under the optimal scheduling solution to Eq.~\ref{eq:overall_goal_seg}, no time utility of the first segment can be increased. 

\noindent 
\textbf{Proof. }
Let $x^*$ be the optimal scheduling solution to Eq.~\ref{eq:overall_goal_seg}. Suppose there exists another solution $x'$ that increases the time utility of the first segment for at least one request $r^j$:
\[
TUF_0^j(W(s_0^j;x'))> TUF_0^j(W(s_0^j;x^*)
\]
We denote the objective value achieved by the scheduling solution $x$ as $v(x)$,

\[
v(x) = \sum_{r^i \in \mathbb{T}} \big(TUF_0^i (W(s_0^i;x)) +\sum_{k=1}^{K^{i}} TUF_1^i (W(s_k^i;x) ) \big )
\]
Since the waiting time of other segments is not impacted by this change (no negative impact on task completion time) and there is no negative impact on other requests, we have:
\[
v(x') > v(x^*)
\]
We have achieved a higher total TUF value with solution $x'$, contradicting optimality. Hence, our initial assumption is false, and it must be that under the optimal solution for Eq.~\ref{eq:overall_goal_seg}, no time utility of the first segment can be further increased without violating optimality.

\noindent \textbf{Conclusion:}  
We have shown that under the optimal solution of Eq.~\ref{eq:overall_goal_seg}, no request completion time can be further reduced and no time utility of the first segment can be increased. These are exactly the two conditions imposed by Eq.~\ref{eq:overall_goal_1}. Hence, the optimal scheduling solution to Eq.~\ref{eq:overall_goal_seg} is Pareto optimal with respect to Eq.~\ref{eq:overall_goal_1}.

\end{document}